\documentclass{article}

\PassOptionsToPackage{numbers,compress}{natbib}

 \usepackage[preprint]{main}

\usepackage[utf8]{inputenc} %
\usepackage[T1]{fontenc}    %
\usepackage[pagebackref=true,breaklinks=true,letterpaper=true,colorlinks,bookmarks=false]{hyperref}
\usepackage{url}            %
\usepackage{booktabs}       %
\usepackage{amsfonts}       %
\usepackage{nicefrac}       %
\usepackage{microtype}      %
\usepackage{amsmath}        %
\usepackage{amssymb}        %
\usepackage{graphicx}
\usepackage{subfig}

\usepackage{multirow}

\usepackage[table,xcdraw]{xcolor}
\definecolor{tabtitle}{gray}{.8}
\definecolor{ours}{gray}{.95}
\definecolor{ggray}{RGB}{127,127,127}
\definecolor{mygray}{RGB}{170,170,170}
\definecolor{mylred}{RGB}{250,130,130}
\definecolor{mylgreen}{RGB}{130,250,130}
\definecolor{mylblue}{RGB}{130,130,250}
\definecolor{reda}{RGB}{202,0,0}
\definecolor{redb}{RGB}{217,148,143}
\definecolor{myyellow}{RGB}{190,144,0}
\definecolor{mygreen}{RGB}{0,136,51}
\definecolor{myblue}{RGB}{0,102,204}
\definecolor{mycongqian}{RGB}{100, 134, 73}

\newcommand{\parhead}[1]{\noindent\textbf{#1}}

\newcommand{\first}[1]{\textcolor{reda}{\textbf{#1}}}
\newcommand{\second}[1]{\textcolor{mygreen}{\textbf{#1}}}
\newcommand{\third}[1]{\textcolor{myblue}{\textbf{#1}}}

\newenvironment{authorfoot}
{}
{}

\newcommand{\myiou}{IoU$_{pix}\uparrow$}
\newcommand{\myniou}{nIoU$_{pix}\uparrow$}
\newcommand{\myfone}{F1$_{pix}\uparrow$}
\newcommand{\myfonetgt}{F1$_{tgt}\uparrow$}
\newcommand{\mypd}{Pd$\uparrow$}
\newcommand{\myfa}{Fa${\times 10^6}\downarrow$}
\newcommand{\myhiou}{hIoU$\uparrow$}

\usepackage{xspace}
\newcommand{\eg}{\emph{e.g.}\xspace} %
\newcommand{\ie}{\emph{i.e.}\xspace} %
\newcommand{\etc}{\emph{etc.}\xspace} %
\newcommand{\vs}{\emph{vs.}\xspace} %

\usepackage{threeparttable}
\usepackage[ruled,vlined,linesnumbered]{algorithm2e}

\SetCommentSty{mycommentfn}
\usepackage{wrapfig2}
\usepackage{enumitem}

\title{Rethinking Evaluation of Infrared Small Target Detection}

\author{%
  Youwei Pang$^{1,3}$~
  Xiaoqi Zhao$^{2}$~
  Lihe Zhang$^{1}$~
  Huchuan Lu$^{1}$ \\
  \textbf{Georges El Fakhri}$^{2}$~
  \textbf{Xiaofeng Liu}$^{2}$~
  \textbf{Shijian Lu}$^{3}$ \\
  $^1$Dalian University of Technology~
  $^2$Yale University~
  $^3$Nanyang Technological University \\
  \texttt{lartpang@gmail.com}
}

\begin{document}

\maketitle
\begin{authorfoot}
    \footnotetext[1]{Corresponding Author: Lihe Zhang (\texttt{zhanglihe@dlut.edu.cn})}
\end{authorfoot}

\begin{abstract}
    As an essential vision task, infrared small target detection (IRSTD)\footnote{In the existing literature, IRSTD can refer to both detection and segmentation tasks. However, \textbf{this paper primarily focuses on the more challenging segmentation task as in~\cite{IRSTD-MSHNet,IRSTD-DNANet,IRSTD-SCTransNet}}.} has seen significant advancements through deep learning.
    However, critical limitations in current evaluation protocols impede further progress.
    First, existing methods rely on fragmented pixel- and target-level specific metrics, which fails to provide a comprehensive view of model capabilities.
    Second, an excessive emphasis on overall performance scores obscures crucial error analysis, which is vital for identifying failure modes and improving real-world system performance.
    Third, the field predominantly adopts dataset-specific training-testing paradigms, hindering the understanding of model robustness and generalization across diverse infrared scenarios.
    This paper addresses these issues by introducing a hybrid-level metric incorporating pixel- and target-level performance, proposing a systematic error analysis method, and emphasizing the importance of cross-dataset evaluation.
    These aim to offer a more thorough and rational hierarchical analysis framework, ultimately fostering the development of more effective and robust IRSTD models.
    An open-source toolkit has be released to facilitate standardized benchmarking.
    \footnote{Our evaluation toolkit: \url{https://github.com/lartpang/PyIRSTDMetrics}}
\end{abstract}

\section{Introduction}
\label{sec:introduction}

Infrared small target detection (IRSTD) is critical in applications such as maritime resource management, navigation, and environmental monitoring~\cite{IRSTD-Maritime,IRSTD-OceanEnvironment,STD-SatelliteImaging,IRSTR-MoCoPnet}.
Infrared imaging leverages the contrast between target and background radiation, offering advantages such as working in all weather conditions and operating day and night.
Detecting small and low-contrast targets in infrared image presents significant challenges due to complex backgrounds, long-distance transmission effects, and a low signal-to-noise ratio~\cite{IRSTDSurvey,IRSTDSurvey-SegmentationNetworks}.
Given the unique challenges, significant research has been devoted to developing algorithms that accurately capture and segment these targets.

Despite advances in detection methods, the evaluation protocols used to benchmark these algorithms remains a subject of concern.
Current IRSTD practices rely on level-specific metrics, including both pixel-level $\text{IoU}_{pix}$, $\text{nIoU}_{pix}$, and $\text{F1}_{pix}$, and target-level Pd and Fa, alongside independent training and testing on individual datasets.
These protocols often prioritize the incomplete evaluation in data-constrained scenarios, failing to deliver comprehensive and detailed model analysis.
Such an inadequate evaluation leads to several issues.
First, existing fragmented and coupled metrics result in a lack of holistic evaluation, making it difficult to fully show the model's true performance.
Furthermore, the current research frequently overlooks systematic failure mode investigation in its rush to report competitive performance benchmarks, which is a fundamental requirement for diagnosing model vulnerabilities and enhancing algorithmic robustness.
Additionally, the evaluation protocol remains constrained by its dataset-specific training-testing paradigm.
Such a widespread practice incentivizes narrow optimization for dataset biases rather than general detection capability.
It not only increases overfitting risks but may also exaggerate perceived performance.

\begin{wrapfigure}[10]{r}{0.5\linewidth}
    \centering
    \includegraphics[width=\linewidth]{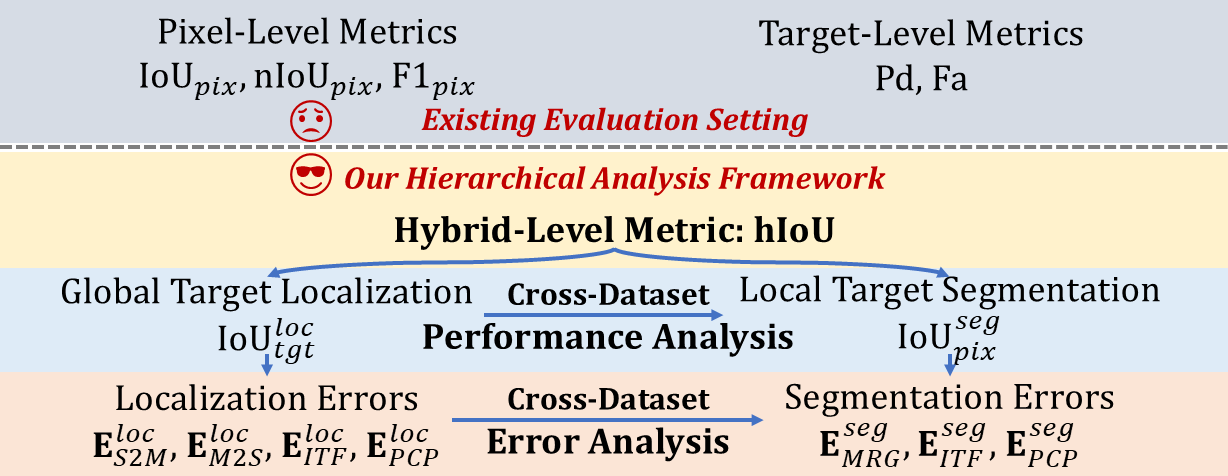}
    \caption{Our hierarchical analysis framework.}
    \label{fig:teaser}
\end{wrapfigure}

To address these challenges, we propose a comprehensive hierarchical analysis framework, which introduces three key components:
1) \textit{a hybrid-level performance metric, hierarchical IoU (hIoU)},
2) \textit{a systematic error analysis method},
and 3) \textit{a cross-dataset evaluation setting}.
Specifically, the proposed hIoU combines the performance from target-level localization and pixel-level segmentation, offering a more holistic view of model efficacy.
And the proposed error analysis method is closely tied and complementary to the proposed metric hIoU.
It allows for a detailed exploration of model failure modes and identifying key cues for improving method effectiveness.
Besides, our cross-dataset setting systematically evaluates model performance across different dataset scenarios, providing valuable measurements of robustness and generalization.
By addressing these challenges, our work provide a thorough and rational evaluation framework, ultimately contributing to the advancement of more reliable and transferable IRSTD applications.

Our main contributions are summarized as follows:
\begin{itemize}[leftmargin=*,itemsep=0em,topsep=0em,parsep=0em]
    \item First to expose limitations in current IRSTD evaluation protocols and propose a hierarchical analysis framework.
    \item Introduce an hybrid-level metric capturing IRSTD performance across target and pixel levels.
    \item Reveal limited cross-dataset generalization of IRSTD algorithms through detailed evaluation.
    \item First to systematically analyze errors in IRSTD by quantifying model limitations under our metric.
    \item Develop a universal and comprehensive evaluation toolkit to advance IRSTD research.
\end{itemize}

\section{Related Work}
\label{sec:related_work}

After the development of several decades, the IRSTD algorithm design has undergone a significant evolution from traditional methods to deep learning techniques~\cite{IRSTDSurvey,IRSTDSurvey-SegmentationNetworks}.
Traditional algorithms~\cite{IRSTD-TraditionalSuppression-NWMTH,IRSTD-TraditionalSuppression-ILCM,IRSTD-TraditionalSuppression-MPCM,IRSTD-TraditionalSuppression-WLDM,IRSTD-TraditionalSuppression-RLCM,IRSTD-TraditionalSuppression-FKRW,IRSTD-TraditionalSuppression-TLLCM,IRSTD-TraditionalSuppression-GSWLCM,IRSTD-TraditionalDecomposition-IPI,IRSTD-TraditionalDecomposition-NIPPS,IRSTD-TraditionalDecomposition-RIPT,IRSTD-TraditionalDecomposition-NRAM,IRSTD-TraditionalDecomposition-NOLC,IRSTD-TraditionalDecomposition-PSTNN} rely on filtering techniques and model-driven approaches, which perform well in simple backgrounds but lacked robustness in complex environments.
Recent advances~\cite{IRSTD-ACM-nIoU,IRSTD-FC3Net,IRSTD-DNANet,IRSTD-ISNet,IRSTD-AGPCNet,IRSTD-UIUNet,IRSTD-RDIAN,IRSTD-MTU-Net,IRSTD-ABC,IRSTD-SeRankDet,IRSTD-MSHNet,IRSTD-MRF3Net,IRSTD-SCTransNet,IRSTD-RPCANet,MIRSTD-DTUM,MIRSTD-RFR,MIRSTD-DeepPro} based on deep learning have significantly propelled IRSTD.

Pioneering work \cite{IRSTD-ACM-nIoU} introduces asymmetric contextual modulation to enhance target-background discrimination while addressing data scarcity via a high-quality annotated dataset.
Subsequent studies focus on specialized architectures.
\cite{IRSTD-DNANet} designs dense nested interaction modules with dual attention to preserve targets across network depths, whereas \cite{IRSTD-FC3Net} proposes feature compensation and cross-level correlation mechanisms to recover lost target details.
For shape-aware detection, \cite{IRSTD-ISNet} integrates Taylor finite difference operators and orientation attention to capture geometric characteristics of targets.
And \cite{IRSTD-MRF3Net} emphasizes lightweight multi-receptive field perception and feature fusion.
Attention mechanisms became pivotal in later innovations.
\cite{IRSTD-AGPCNet} developes pyramid context modules with global-local attention for complex backgrounds, while \cite{IRSTD-UIUNet} embeds nested U-Nets with resolution-maintenance supervision to enhance multi-scale contrast.
And \cite{IRSTD-SeRankDet} employs selective rank-aware attention to resolve hit-miss trade-offs.
Transformer-based architectures emerge as powerful alternatives.
\cite{IRSTD-ABC} combines bilinear correlation and dilated convolutions for semantic refinement.
\cite{IRSTD-SCTransNet} introduces spatial-channel cross-transformers to model full-level semantic differences.
Besides, the loss function design proves critical.
Scale- and location-sensitive losses \cite{IRSTD-MSHNet} are developed to address target scale and location variations.
Efforts to balance performance and interpretability are also explored by \cite{IRSTD-RPCANet} which unfolds robust PCA into a deep network.

Despite these algorithmic advancements, the field has predominantly focused on architectural innovations while neglecting improvements in evaluation frameworks.
Our work shifts the focus from algorithm design to evaluation pipeline.
And our framework provides actionable insights for real-world deployments, complementing rather than competing with existing algorithmic advancements.

\section{Evaluation Metrics}

\subsection{Preliminaries}

The existing deep learning-based IRSTD approaches utilize the hierarchical encoder-decoder\cite{Unet,FCN} architecture to process a dataset with $K$ pairs of thermal infrared images $\{I_i\}_{i=1}^{K}$ and the corresponding ground truth (GT) binary masks $\{G_i\}_{i=1}^{K}$.
In the pipeline, gray-scale predictions $\{P_i\}_{i=1}^{K}$ will be generated for these input images by the deep model in an end-to-end manner.
The pixel values of $P_i$ represent the probabilities of pixels belonging to infrared small targets within the $i^{th}$ input image, and they range from $0$ to $1$.
$0$ indicates that the pixel is classified as a background pixel, whereas the value $1$ signifies that the pixel is considered to be part of the foreground target.
Gray-scale predictions $\{P_i\}_{i=1}^{K}$ are directly compared with their corresponding binary GT masks $\{G_i\}_{i=1}^{K}$ to calculate similarity.
Thus, the average performance on the dataset can be calculated to evaluate the proposed algorithm.
Currently, IRSTD metrics can be broadly classified into pixel-level and target-level categories according to their computational primitives.
The following sections will provide detailed introductions.

\subsection{Pixel-level Metrics}
\label{sec:pixel_level_metrics}

In this branch, existing metrics are all computed based on the statistical values the number of pixel-level
true positives (TP$_{pix}$),
false positives (FP$_{pix}$),
true negatives (TN$_{pix}$), %
and false negatives (FN$_{pix}$)
from the binary confusion matrix.
Note that TP$_{pix}$, FP$_{pix}$, TN$_{pix}$, and FN$_{pix}$ rely on the binary predictions and GTs with the same size.
Therefore, the binarization strategy applied to the gray-scale predictions also influences the results of these metrics.
However, the strategy design is beyond the scope of this work.
Unless otherwise specified, a commonly-used threshold of $0.5$ will be used by default for the prediction thresholding.
The pixel-level metrics involved in existing works include Intersection over Union (IoU), and F1-score (F1)
\footnote{While IoU and F1 fundamentally measure region overlap, their standard implementation in IRSTD prioritizes global pixel-level evaluation over target-level alignment.}.

\parhead{Intersection over Union (IoU)} is a widely-used metric for measuring the overlap between the predicted foreground and the GT mask, normalizing their intersection cardinality with respect to the union.
In existing works, there exist two different variants according to the difference of computational logic, including the conventional IoU ($\text{IoU}_{pix}$) and the normalized IoU ($\text{nIoU}_{pix}$)~\cite{IRSTD-ACM-nIoU} as follows:
\begin{align}
    \text{IoU}_{pix}  = \frac{ \sum_{i=1}^{K} |\text{TP}_{pix}^{[i]}| }{ \sum_{i=1}^{K} (|\text{TP}_{pix}^{[i]}| + |\text{FP}_{pix}^{[i]}| + |\text{FN}_{pix}^{[i]}|) } \quad
    \text{nIoU}_{pix} = \sum_{i=1}^{K} \frac{|\text{TP}_{pix}^{[i]}| / {K}}{|\text{TP}_{pix}^{[i]}| + |\text{FP}_{pix}^{[i]}| + |\text{FN}_{pix}^{[i]}|}
    \label{equ:iou}
\end{align}
where $|\cdot|$ and $[i]$ represent the number of elements in the set and the sample index, respectively.
$\text{IoU}_{pix}$ aggregates global pixel statistics across all samples.
$\text{nIoU}_{pix}$ computes per-sample IoU before averaging, which ensures equal contribution from all samples, thereby fairly evaluating performance across diverse targets.
When there is only one single sample (\ie, $K=1$), $\text{nIoU}_{pix} = \text{IoU}_{pix}$.

\parhead{F1-score (F1$_{pix}$)} is the harmonic mean of precision ($\text{Pre}_{pix}$) and recall ($\text{Rec}_{pix}$), designed to balance these metrics and provide a comprehensive evaluation.
The calculation framework is defined as:
\begin{gather}
    \label{equ:f_measure}
    \text{Pre}_{pix} = \frac{\sum_{i=1}^{K} |\text{TP}_{pix}^{[i]}|}{\sum_{i=1}^{K} |\text{TP}_{pix}^{[i]}| + |\text{FP}_{pix}^{[i]}|} \quad
    \text{Rec}_{pix} = \frac{\sum_{i=1}^{K} |\text{TP}_{pix}^{[i]}|}{\sum_{i=1}^{K} |\text{TP}_{pix}^{[i]}| + |\text{FN}_{pix}^{[i]}|} \\
    \text{F1}_{pix}  = \frac{2 \text{Pre}_{pix} \times \text{Rec}_{pix}}{\text{Pre}_{pix} + \text{Rec}_{pix}}
\end{gather}
where $\text{Pre}_{pix}$ and $\text{Rec}_{pix}$ are primarily used to calculate the metric F1 and not used independently.

\subsection{Target-level Metrics}
\label{sec:target_level_metrics}

Target-level metrics are designed to evaluate model performance at the level of individual targets and play an important role in IRSTD~\cite{IRSTD-DNANet}.
Unlike pixel-level metrics that aggregate performance across the entire image, target-level metrics focus on the localization quality of predictions for each distinct small target.
And target region sets $\{T_G^m\}$ and $\{T_P^n\}$ are extracted by the connected component analysis algorithm from the GT masks and binary predictions, respectively.
Each individual predicted target $T_P^n$ tries to match a nearby target $T_G^m$ in the GT mask based on a specific criteria.
Existing works~\cite{IRSTD-DNANet,IRSTD-ABC,IRSTD-MTU-Net} usually use \textit{the centroid distance} to determine the matching relationship.
If the distance between $T_P^n$ and $T_G^m$ is less than the predefined threshold (\eg, 3~\cite{IRSTD-DNANet}), they will be viewed as the matched pair.
Every predicted target matches to at most one GT target, and vice versa.
Eventually, the predicted target regions matched to GT targets can be considered as belonging to target-level TP (TP$_{tgt}$), while the remaining $T_P$ and $T_G$ belong to FP$_{tgt}$ and FN$_{tgt}$, respectively.

\parhead{Probability of Detection (Pd).}
It quantifies the model's ability to detect true targets by computing the fraction of correctly matched target predictions (the count of TP$_{tgt}$) over the total number of GT targets (TP$_{tgt}$ + FN$_{tgt}$) as follows:
\begin{gather}
    \text{Pd} = \frac{ \sum_{i=1}^{K} |\text{TP}_{tgt}^{[i]}| }{ \sum_{i=1}^{K} (|\text{TP}_{tgt}^{[i]}| + |\text{FN}_{tgt}^{[i]}|) }
\end{gather}
It effectively evaluates detection completeness for small targets regardless of their pixel area size.

\parhead{False-Alarm Rate (Fa).}
It evaluates the spatial impact of unmatched predictions by normalizing the count of pixels in all FP$_{tgt}$ targets against the entire image resolution:
\begin{align}
    \text{Fa} = \frac{ \sum_{i=1}^{K} \texttt{TargetAreaOf}(\text{FP}_{tgt}^{[i]}) }{ \sum_{i=1}^{K} \texttt{ImageAreaOf}(G_i) }
\end{align}
where the normalization stabilizes comparisons across varying resolutions.

\begin{figure*}[!t]
    \centering
    \subfloat[]{\centering
        \label{fig:metric_comparison_1}
        \includegraphics[width=0.16\linewidth]{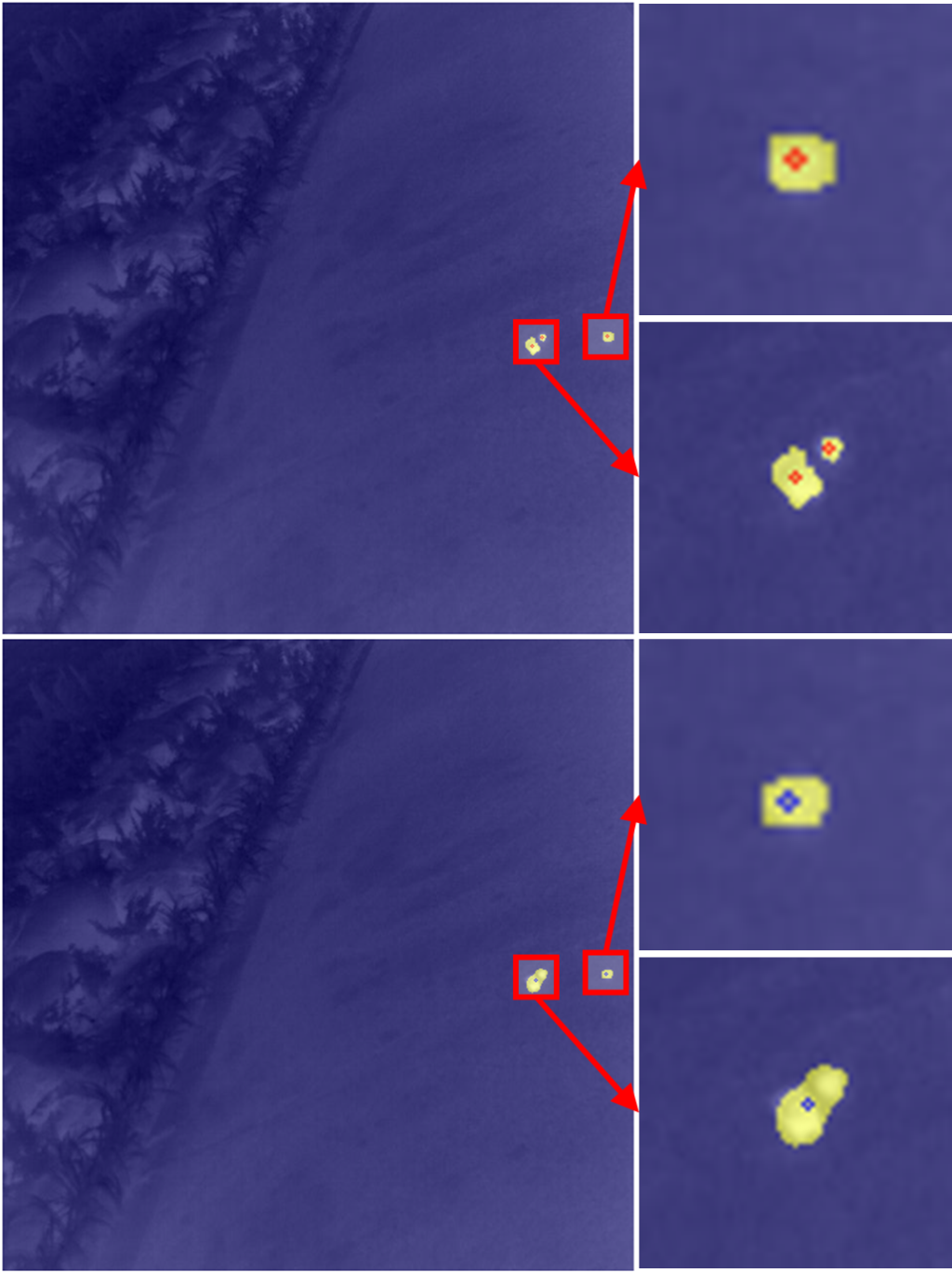}}
    \subfloat[]{\centering
        \label{fig:metric_comparison_2}
        \includegraphics[width=0.16\linewidth]{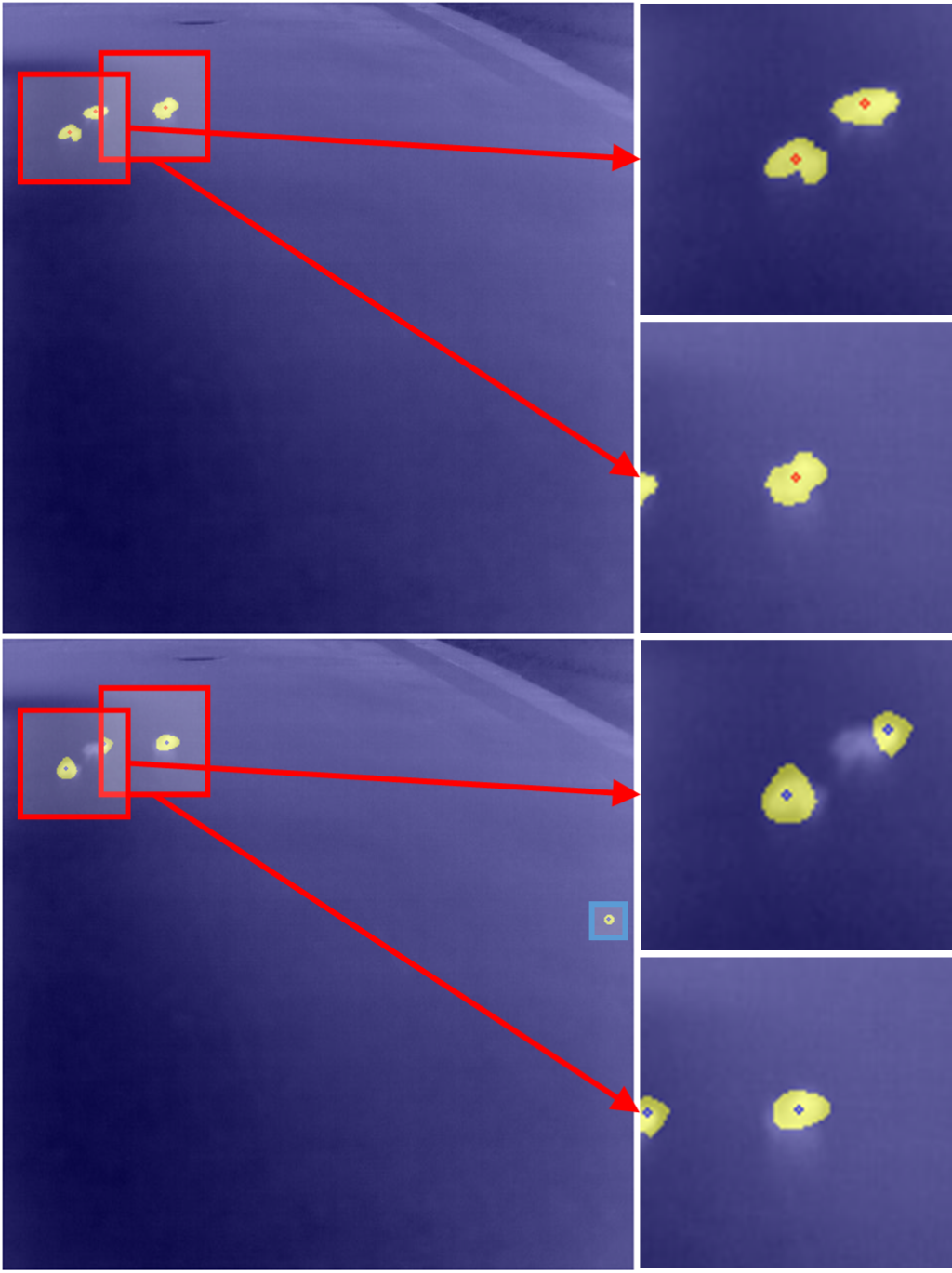}}
    \subfloat[]{\centering
        \label{fig:metric_comparison_3}
        \includegraphics[width=0.16\linewidth]{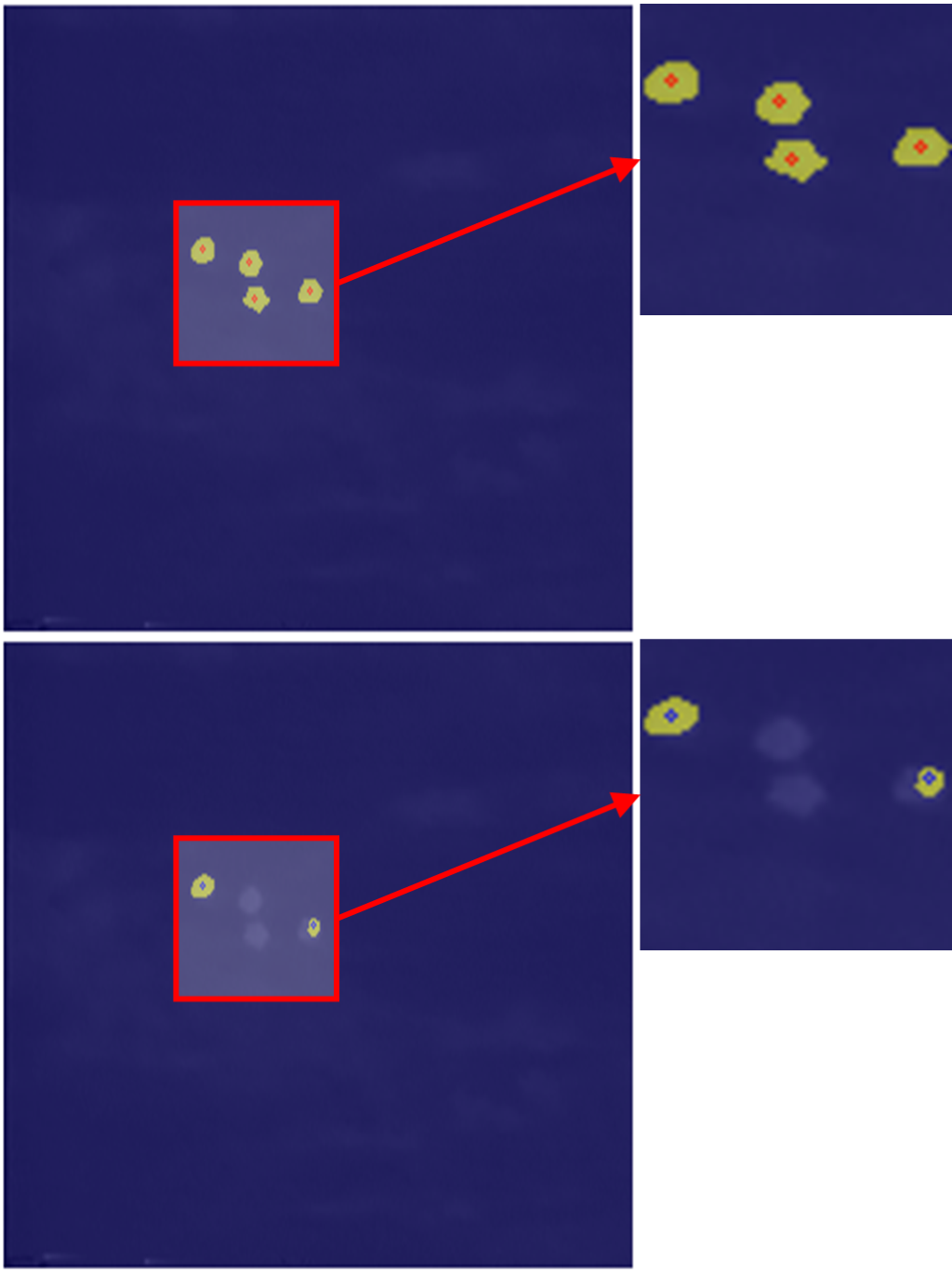}}
    \subfloat[]{\centering
        \label{fig:metric_comparison_4}
        \includegraphics[width=0.16\linewidth]{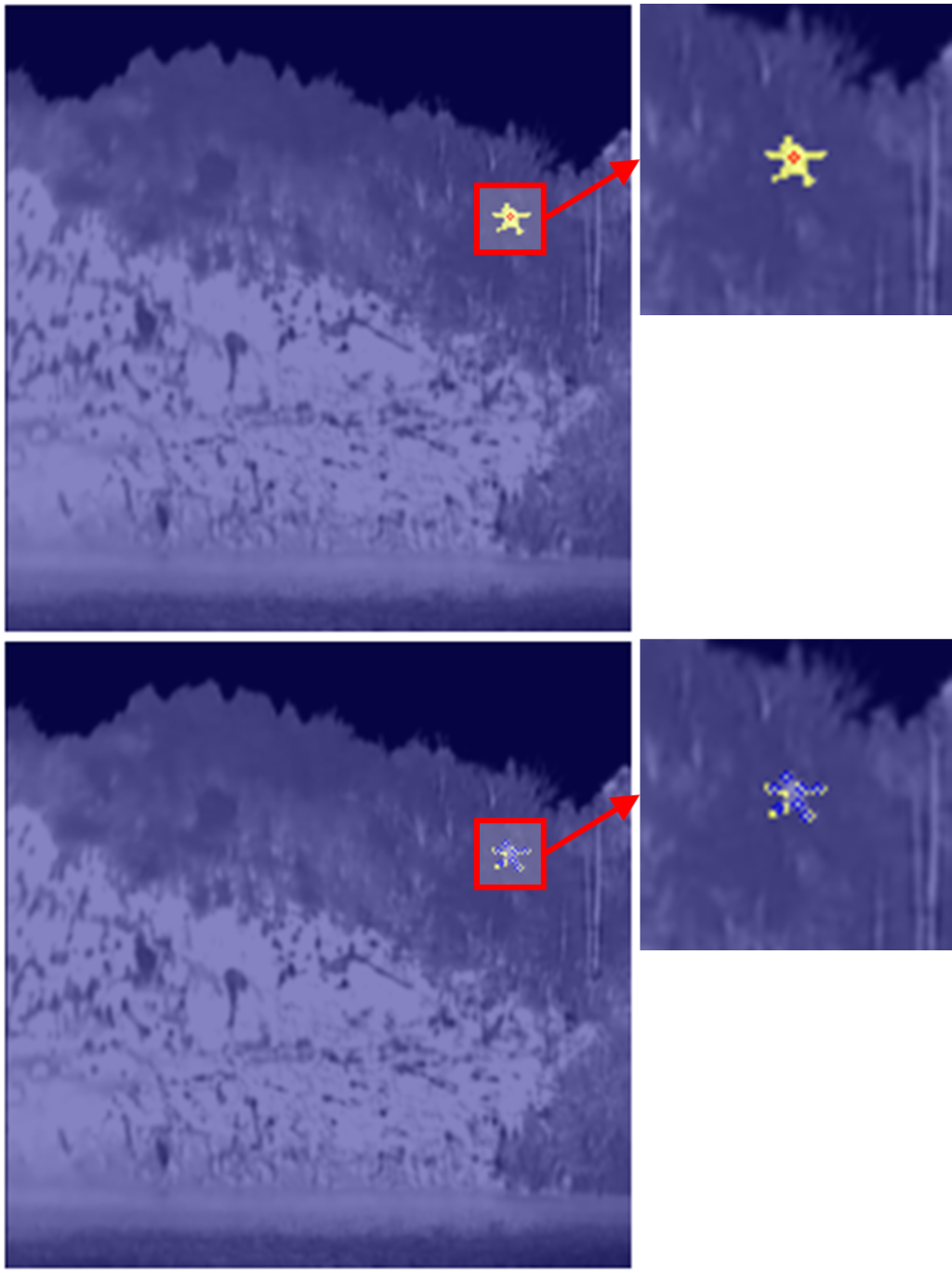}}
    \subfloat[]{\centering
        \label{fig:metric_comparison_5}
        \includegraphics[width=0.16\linewidth]{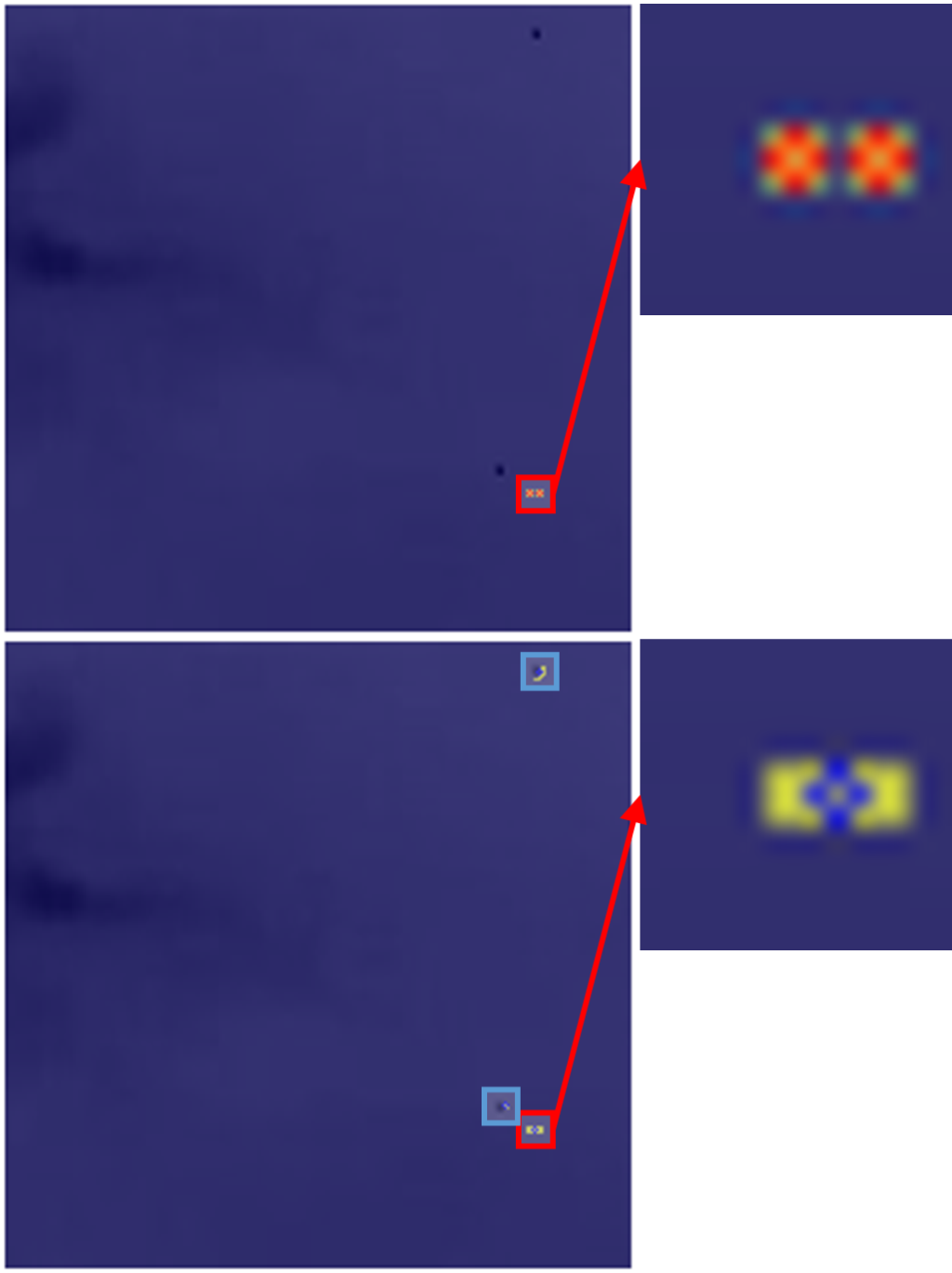}}
    \subfloat[]{\centering
        \label{fig:metric_comparison_6}
        \includegraphics[width=0.16\linewidth]{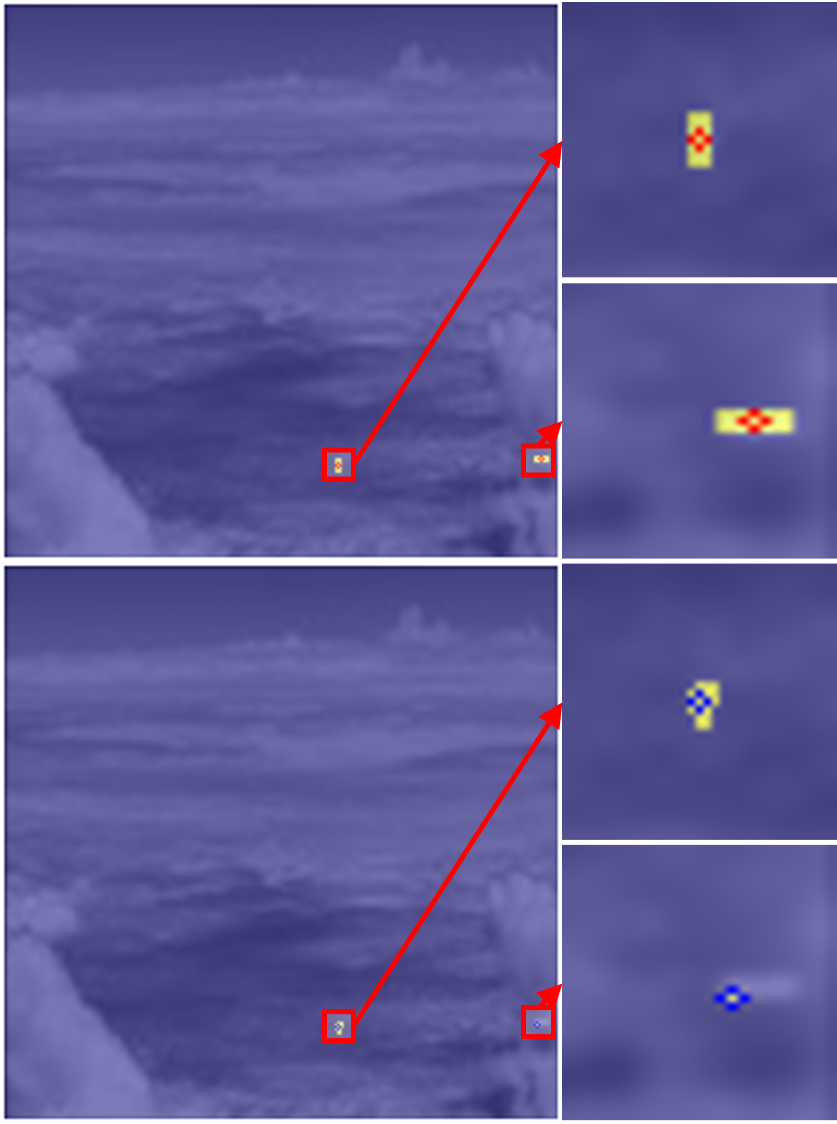}}
    \par
    \vspace{-2ex}
    \subfloat{\centering
        \resizebox{0.8\linewidth}{!}{
\begin{tabular}{c|rrrrrrr|r|rrrr|rrrrr}
    \toprule[2pt]
                                     & \myiou & \myniou & \myfone & \multicolumn{2}{c}{\mypd} & \multicolumn{2}{c|}{\myfa} & \myhiou                  & IoU$^{seg}_{pix}\uparrow$ & $\mathbf{E}^{seg}_{MRG}\downarrow$ & $\mathbf{E}^{seg}_{ITF}\downarrow$ & $\mathbf{E}^{seg}_{PCP}\downarrow$ & IoU$^{loc}_{tgt}\uparrow$ & $\mathbf{E}^{loc}_{S2M}\downarrow$ & $\mathbf{E}^{loc}_{M2S}\downarrow$ & $\mathbf{E}^{loc}_{ITF}\downarrow$ & $\mathbf{E}^{loc}_{PCP}\downarrow$                 \\
    \multirow{-2}{*}{Subfigure}      &        &         &         & Ori.~\cite{IRSTD-DNANet}  & OPDC                       & Ori.~\cite{IRSTD-DNANet} & OPDC                      & OPDC                               & OPDC                               & OPDC                               & OPDC                      & OPDC                               & OPDC                               & OPDC                               & OPDC                               & OPDC  & OPDC  \\
    \midrule[1pt]
    \subref{fig:metric_comparison_1} & 0.573  & 0.573   & 0.728   & 0.333                     & 0.667                      & 785.828                  & 0.000                     & 0.373                              & 0.560                              & 0.052                              & 0.246                     & 0.142                              & 0.667                              & 0.333                              & 0.000                              & 0.000 & 0.000 \\
    \subref{fig:metric_comparison_2} & 0.438  & 0.438   & 0.610   & 0.333                     & 1.000                      & 1457.214                 & 198.364                   & 0.344                              & 0.459                              & 0.000                              & 0.225                     & 0.316                              & 0.750                              & 0.000                              & 0.000                              & 0.250 & 0.000 \\
    \subref{fig:metric_comparison_3} & 0.310  & 0.310   & 0.473   & 0.500                     & 0.500                      & 0.000                    & 0.000                     & 0.300                              & 0.600                              & 0.000                              & 0.000                     & 0.400                              & 0.500                              & 0.000                              & 0.000                              & 0.000 & 0.500 \\
    \subref{fig:metric_comparison_4} & 0.477  & 0.477   & 0.646   & 1.000                     & 1.000                      & 579.834                  & 625.610                   & 0.005                              & 0.036                              & 0.000                              & 0.000                     & 0.964                              & 0.143                              & 0.000                              & 0.857                              & 0.000 & 0.000 \\
    \subref{fig:metric_comparison_5} & 0.514  & 0.514   & 0.679   & 0.500                     & 0.500                      & 244.141                  & 244.141                   & 0.118                              & 0.474                              & 0.474                              & 0.053                     & 0.000                              & 0.250                              & 0.250                              & 0.000                              & 0.500 & 0.000 \\
    \subref{fig:metric_comparison_6} & 0.326  & 0.326   & 0.492   & 1.000                     & 1.000                      & 0.000                    & 0.000                     & 0.313                              & 0.313                              & 0.000                              & 0.085                     & 0.602                              & 1.000                              & 0.000                              & 0.000                              & 0.000 & 0.000 \\
    \bottomrule[2pt]
\end{tabular}

}}
    \caption{Comparison of different metrics.
        \textcolor{red}{Red} and \textcolor{blue}{blue} boxes to highlight the target regions.
        \textcolor{red}{Red} and \textcolor{blue}{blue} points indicate the target centroids in ground truth (GT) masks and predictions, respectively.
        Zoom in on the digital color version for details.
        ``Ori.''~\cite{IRSTD-DNANet} and ``OPDC'' refer to the original distance-based strategy and the proposed OPDC strategy for target matching.
    }
    \label{fig:metric_comparison}
    \vspace{-1ex}
\end{figure*}

\subsection{Current Limitations}
\label{sec:current_limitations}

\parhead{Pixel-level Evaluation.}
Although IoU$_{pix}$, nIoU$_{pix}$, and F1$_{pix}$ mitigate foreground imbalance by incorporating relative area proportions, they are inherently limited to global pixel-wise analysis and cannot assess target-level spatial localization or segmentation accuracy, both of which are critical to understanding the fine-grained IRSTD performance.

\parhead{Target-level Evaluation.}
Pd measures detection completeness but ignores false positives, potentially overestimating performance in noisy environments.
Fa quantifies the spatial impact of FP$_{tgt}$ by normalizing their total area to the full image.
But it maybe undercount smaller FP$_{tgt}$ targets while overpenalizing large FP$_{tgt}$ targets.
Additionally, Fa disregards pixel-level errors (FP$_{pix}$) within TP$_{tgt}$ targets and target-level errors FN$_{tgt}$ targets.
Although conventional Pd and Fa may compensate for each other's blind spots, their shared dependency on the distance threshold and data biases related to size and shape limit their ability to holistically reflect algorithm performance.

\parhead{Fragmented Evaluation Paradigm.}
Current pipelines rely on several pixel-level (IoU$_{pix}$, nIoU$_{pix}$, and F1$_{pix}$) and target-level (Pd and Fa) metrics to approximate holistic performance.
The former lacks spatial awareness, while the latter oversimplifies error patterns.
While they individually address specific aspects, their simple combination creates a mismatch between evaluation pipeline and task requirements.
Such fragmented paradigm obscures critical trade-offs, such as segmentation-localization dependency and target diversity tolerance, leaving incomplete or even contradictory performance insights.
A new evaluation framework, considering hybrid-level modeling and data diversity, is essential to overcome these limitations.

\section{New Evaluation Framework}
\label{sec:new_evaluation_framework}

\subsection{Target Matching Strategy}
\label{sec:our_matching}

Current target-level IRSTD metrics suffer from overly strict distance-only filtering~\cite{IRSTD-DNANet}, where centroid matching frequently misjudges offset, fragmented, or connected predictions as shown in Fig.~\ref{fig:metric_comparison_1} and Fig.~\ref{fig:metric_comparison_2}.
By introducing overlap-priority constraint to enhance the matching mechanism, we propose the ``Overlap Priority with Distance Compensation'' (OPDC) strategy (Alg.~\ref{alg:matching}), which can effectively alleviates these limitations.

\parhead{Overlap priority constraint} enforces shape coherence by computing pairwise overlap ratios between targets from GTs and predictions.
For each target pair, if their IoU exceeds 0.5, it is marked as a valid candidate.
The commonly-used assignment algorithm~\cite{linear_sum_assignment} is then applied to the full centroid Euclidean distance matrix to find the minimum-cost matching, followed by retaining only valid pairs satisfying the overlap constraint.
This phase ensures morphological alignment by filtering mismatches caused by unreasonable shapes or over-prediction as the smallest target in Fig.~\ref{fig:metric_comparison_1}, resulting in initial matched pairs $S_{TP}$ and unmatched sets $S_{FN}/S_{FP}$.

\parhead{Distance-based compensation} supplements residual unmatched pairs by evaluating centroid Euclidean distances.
For targets in $S_{FN}$ and $S_{FP}$, pairs with distances below the strict threshold (3 pixels, as in~\cite{IRSTD-DNANet}) are re-matched via the assignment algorithm~\cite{linear_sum_assignment}.
This phase specifically addresses small or low-overlap targets where shape metrics may underperform, leveraging spatial proximity as a secondary criterion to reassess their value.

By hierarchically integrating overlap and distance constraints, OPDC achieves a more intuitive matching.
The former prioritizes overlap-based filtering to relax the original distance constraint, aligning with the real-world intuition where high overlap inherently indicates true morphological correspondence.
The latter acts as a safety net exclusively for low-overlap residuals, ensuring spatial proximity without compromising the dominance of shape-aware matching.
As shown in Fig.~\ref{fig:metric_comparison_6}, the right predicted target that do not satisfy the overlap constraint are re-matched with the GT target in the distance compensation.

\subsection{Hierarchical Intersection over Union}
\label{sec:hiou}

Evaluation practices in current IRSTD studies typically focus on either isolated pixel-level or target-level similarity measurements between predictions and GTs.
However, we propose a new hybrid-level metric, \ie, hierarchical Intersection over Union (hIoU), which hierarchically combines both global target-level localization and local pixel-level segmentation performance as follows:
\begin{gather}
    \label{equ:locsegiou}
    \text{IoU}^{loc}_{tgt} = \frac{ \sum_{i=1}^{K} |\text{TP}_{tgt}^{[i]}| }{ \sum_{i=1}^{K} |\text{TP}_{tgt}^{[i]}| + |\text{FP}_{tgt}^{[i]}| + |\text{FN}_{tgt}^{[i]}| }     \quad
    \text{IoU}^{seg}_{pix} = \frac{\sum_{(T_{G}^m, T_{P}^n) \in \text{TP}_{tgt}} {(T_{G}^m \cap T_{P}^n)}/{(T_{G}^m \cup T_{P}^n)}}{ \sum_{i=1}^{K} |\text{TP}_{tgt}^{[i]}|} \\
    \label{equ:hiou}
    \text{hIoU}            = \text{IoU}^{loc}_{tgt} \times \text{IoU}^{seg}_{pix} %
\end{gather}
where ${ (T_{G}^m \cap T_{P}^n) } / { (T_{G}^m \cup T_{P}^n) }$ denotes the intersection-over-union ratio of the matched predicted target $T_{P}^n$ and GT target $T_{G}^m$ from $\text{TP}_{tgt}$.
$\text{IoU}^{loc}_{tgt}$ and $\text{IoU}^{seg}_{pix}$ reflect the IoU-based localization and segmentation performance in infrared small target prediction, respectively.
Specifically, we first adjust the IoU metric to measure the target-level localization performance $\text{IoU}^{loc}_{tgt}$.
And then, for those predicted targets matched with GT targets, $\text{IoU}^{seg}_{pix}$ is further applied to measure the local fine-grained segmentation similarity between them and the corresponding GT targets.
The multiplicative combination in hIoU inherently balances localization and segmentation: a method missing targets (low $\text{IoU}_{tgt}^{loc}$) or producing imprecise regions (low $\text{IoU}_{pix}^{seg}$) will be penalized proportionally.
Unlike additive combinations that allow for linear error compensation (\eg, high localization scores masking poor segmentation), this coupling measures the joint performance in the unit space $[0, 1]^2$, which makes improvements in one part necessary to maintain the other's performance.
This hybrid-level paradigm enables more comprehensive and nuanced performance evaluation, particularly crucial for the IRSTD task where \textit{both complete localization (preventing target loss)} and \textit{precise region delineation (ensuring target integrity)} are equally valuable for practical applications.
Further discussions can be found in Sec.~\ref{supp:why_use_multiplicative_form}.

\subsection{Error Analysis Method}
\label{sec:error_analysis_method}

\begin{wrapfigure}[16]{r}{0.5\linewidth}
    \centering
    \includegraphics[width=0.8\linewidth]{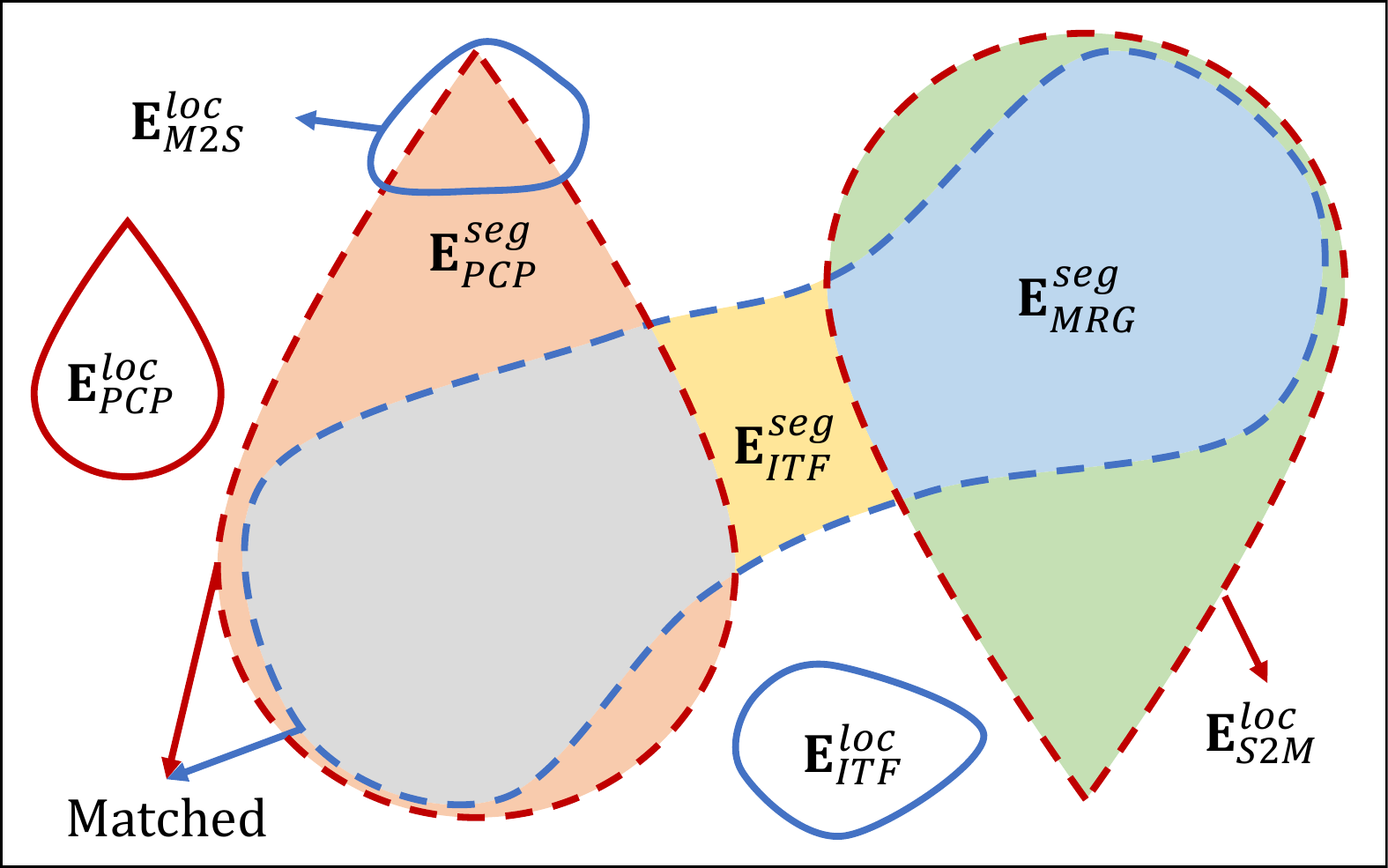}
    \caption{
        Error types (Sec.~\ref{sec:error_analysis_method}) for three predicted and three GT targets.
        \textcolor{blue}{Blue} contours denote predictions and \textcolor{red}{red} contours denote GT.
        Under our OPDC strategy, only the middle prediction is matched to a GT and all others remain unmatched.
    }
    \label{fig:error_analysis}
\end{wrapfigure}

Existing protocols for evaluating IRSTD typically focus on overall average performance values.
This analytical preference obscures critical failure modes and makes it difficult to diagnose and improve model deficiencies.
For instance, a low $\text{IoU}_{pix}$ and Pd could stem from background noise interference, adjacent target merging, or  target perception limitations---each requiring distinct corrective strategies.
To address this, building on our hierarchical evaluation paradigm as stated in Sec.~\ref{sec:hiou}, we categorize prediction errors into two associated levels:
\textbf{target-level localization errors} ($\textbf{E}^{loc}_{S2M}$, $\textbf{E}^{loc}_{M2S}$, $\textbf{E}^{loc}_{ITF}$, and $\textbf{E}^{loc}_{PCP}$) and
\textbf{pixel-level segmentation errors} ($\textbf{E}^{seg}_{MRG}$, $\textbf{E}^{seg}_{ITF}$, and $\textbf{E}^{seg}_{PCP}$).
They provide fine-grained decomposition of the performance losses reflected in $\text{IoU}^{\textit{loc}}_{\textit{tgt}}$ (target localization IoU) and $\text{IoU}^{\textit{seg}}_{\textit{pix}}$ (target segmentation IoU) metrics, respectively.
Critically, the total error for each level is complementary to its corresponding IoU metric:
\begin{align}
    \label{equ:total_error}
    \text{Total Localization Error } \mathbf{E}^{loc} = 1 - \text{IoU}^{loc}_{tgt} \quad
    \text{Total Segmentation Error } \mathbf{E}^{seg} = 1 - \text{IoU}^{seg}_{pix}
\end{align}
Such a design establishes an explicit error-accuracy complementary relationship, where our error subtypes illustrated in Fig.~\ref{fig:error_analysis} quantify distinct sources of deviation from the ideal performance of 1.

\parhead{Target-level localization errors}
quantify mismatches and misidentifications in perceiving individual targets, including $\textbf{E}^{loc}_{S2M}$, $\mathbf{E}^{loc}_{M2S}$, $\mathbf{E}^{loc}_{ITF}$, and $\mathbf{E}^{loc}_{PCP}$.
\begin{itemize}[leftmargin=*,itemsep=0em,topsep=0em,parsep=0em]
    \item $\textbf{E}^{loc}_{S2M}$ (Single-to-Multi Mismatch):
          During the matching process, when a single predicted target satisfies overlap/distance constraints with multiple GT targets but ultimately gets assigned to only one GT target, the remaining unmatched GT targets contribute to this localization error.
          This is commonly observed in dense target clusters where a single prediction overlaps ambiguously with multiple GT targets, as illustrated in Fig.~\ref{fig:metric_comparison_1} and Fig.~\ref{fig:metric_comparison_5}.
    \item $\mathbf{E}^{loc}_{M2S}$ (Multi-to-Single Mismatch):
          When multiple predicted targets satisfy matching constraints with a single GT target, the one-to-one matching protocol forces retention of only one prediction.
          The excluded and unmatched competing predictions then become contributors to this localization error.
          This reflects model oversensitivity or fragmented predictions near valid targets, as demonstrated in Fig.~\ref{fig:metric_comparison_4}.
    \item $\mathbf{E}^{loc}_{ITF}$ (Interference Error):
          It corresponds to other false target predictions with no corresponding GT, which are primarily triggered by background clutter or noise artifacts, as marked by blue boxes in Fig.~\ref{fig:metric_comparison_2} and Fig.~\ref{fig:metric_comparison_5}.
    \item $\mathbf{E}^{loc}_{PCP}$ (Perception Error):
          It quantifies undetected GTs that fail to meet matching criteria, as exemplified in Fig.~\ref{fig:metric_comparison_3}.
          This indicates model insensitivity to low-contrast or morphologically variable targets.
\end{itemize}

\parhead{Pixel-level segmentation errors}
capture neighboring interferences and shape distortions in matched targets, including $\textbf{E}^{seg}_{MRG}$, $\mathbf{E}^{seg}_{ITF}$, and $\mathbf{E}^{seg}_{PCP}$.
\begin{itemize}[leftmargin=*,itemsep=0em,topsep=0em,parsep=0em]
    \item $\textbf{E}^{seg}_{MRG}$ (Merging Error):
          This error quantifies false positive pixels within the regions of the GT targets that are not matched to the current predicted target.
          It specifically occurs when a target prediction extends into neighboring GT target regions, particularly when adjacent targets are erroneously merged, as shown in Fig.~\ref{fig:metric_comparison_1} and Fig.~\ref{fig:metric_comparison_5}.
    \item $\mathbf{E}^{seg}_{ITF}$ (Interference Error):
          It represents incorrectly predicted foreground pixels in real background region within target neighborhood, highlighting local noise interference or incomplete background suppression, as exemplified in Fig.~\ref{fig:metric_comparison_1} and Fig.~\ref{fig:metric_comparison_2}.
    \item $\mathbf{E}^{seg}_{PCP}$ (Perception Error):
          It accounts for the missing prediction within the region of the matched GT target, reflecting model uncertainty in delineating targets with faint edges or complex structures, as illustrated in Fig.~\ref{fig:metric_comparison_4}.
\end{itemize}

\section{Experiment}

\subsection{Experimental Setup}
\label{sec:experimental_setup}

\parhead{Datasets.}
We evaluate IRSTD methods on three widely-used datasets: IRSTD1k~\cite{IRSTD-ISNet}, SIRST~\cite{IRSTD-ACM-nIoU}, and NUDT~\cite{IRSTD-DNANet}.
These datasets exhibit inherent diversity and complexity~\cite{IRSTD-SeRankDet}, enabling comprehensive evaluation of both within-dataset performance and cross-dataset generalization.

\parhead{Metrics.}
We adopt several existing standard metrics, \ie, $\text{IoU}_{pix}$, $\text{nIoU}_{pix}$, $\text{F1}_{pix}$, Pd, and Fa.
Besides, we introduce the proposed hIoU to holistically evaluate localization-segmentation joint performance.
For error analysis, we decompose hIoU into two components, \ie, $\text{IoU}^{loc}_{tgt}$ and $\text{IoU}^{seg}_{pix}$, and further break down their error statistics into fine-grained aspects based on a set of error subtypes.

\parhead{Implementation Details.}
To ensure fair comparison across methods, we retrain all approaches using their official codebases under strictly controlled settings.
All models are optimized via Adam optimizer with an initial learning rate of 0.0005 and a multi-step decay schedule.
They are trained for 400 epochs using a batch size of 16 on two NVIDIA GeForce RTX 3090 GPUs with a total of 48 GB memory.
Following the existing practices~\cite{IRSTD-MSHNet}, we introduce random horizontal flipping, cropping, and blurring to mitigate the overfitting risk.
Besides, all images are bilinearly interpolated to 256 $\times$ 256 resolution during both training and testing phases.

\subsection{Results and Discussion}
\label{sec:results_discussion}

We conduct a comprehensive benchmarking study of 14 recent deep learning-based\footnote{Deep learning approaches are prioritized due to their demonstrated superiority in handling the IRSTD task.}
IRSTD methods~\cite{IRSTD-ACM-nIoU,IRSTD-FC3Net,IRSTD-DNANet,IRSTD-ISNet,IRSTD-AGPCNet,IRSTD-UIUNet,IRSTD-RDIAN,IRSTD-MTU-Net,IRSTD-ABC,IRSTD-SeRankDet,IRSTD-MSHNet,IRSTD-MRF3Net,IRSTD-SCTransNet,IRSTD-RPCANet} through and dual evaluation protocols: conventional metrics and our hierarchical analysis framework.
This analysis uniquely addresses two critical gaps in existing research:
1) over-optimistic single-dataset evaluation and
2) lack of detailed error analysis.
By introducing cross-dataset setting and fine-grained error decomposition, we reveal previously obscured performance limitations.

\parhead{Holistic Performance.}
Existing metrics show weak alignment in practical performance evaluation.
As exemplified in Tab.~\ref{tab:metric_all_datasets}, MSHNet~\cite{IRSTD-MSHNet}, achieving top scores in conventional metrics (IoU$_{pix}$, F1$_{pix}$ and Fa), fails to translate these advantages into superior holistic performance (hIoU).
While DNANet~\cite{IRSTD-DNANet} does not stand out in conventional metrics, its more balanced segmentation and localization performance propels it to hIoU leadership (0.557 \vs MSHNet's 0.549), as shown in Tab.~\ref{tab:error_datasets}.
This inconsistency stems from their narrow focus on isolated aspects (pixel-level segmentation accuracy, target-level detection recall, \etc.) without adequately modeling task hierarchy.
Our hIoU demonstrates important value by reconciling these different aspects.

\parhead{OPDC-Based Matching.}
To show OPDC's effects, we recalculated Pd and Fa metrics under this strategy (``+OPDC'') as listed in Tab.~\ref{tab:metric_all_datasets}.
The observed recall improvement reveals that the original distance-based criterion overly constrained matching, discarding predictions with valuable positional reference for target localization.
As shown in Fig.~\ref{fig:metric_comparison}, when predictions span multiple adjacent GT targets (Fig.~\ref{fig:metric_comparison_1} and Fig.~\ref{fig:metric_comparison_5}) or partially intersect isolated targets (Fig.~\ref{fig:metric_comparison_2}), they fail centroid distance checks under the current protocol, but provide key location cues.
OPDC alleviates these limitations through a dual-constraint mechanism that prioritizes overlap constraint as the primary matching trigger, with centroid distance constraint acting as supplementary validators as stated in Sec.~\ref{sec:our_matching}.

\begin{table*}[!t]
    \centering
    \caption{Cross-dataset performance analysis. Colors \first{red}, \second{green} and \third{blue} represent the first, second and third ranked results.}
    \label{tab:metric_all_datasets}
    \resizebox{\linewidth}{!}{
\begin{tabular}{ll|rrrrrrrrrrrrrr}
    \toprule[2pt]
                                 &
                                 & ACM$_{21}$~\cite{IRSTD-ACM-nIoU}
                                 & FC3Net$_{22}$~\cite{IRSTD-FC3Net}
                                 & DNANet$_{22}$~\cite{IRSTD-DNANet}
                                 & ISNet$_{22}$~\cite{IRSTD-ISNet}
                                 & AGPCNet$_{23}$~\cite{IRSTD-AGPCNet}
                                 & UIUNet$_{23}$~\cite{IRSTD-UIUNet}
                                 & RDIAN$_{23}$~\cite{IRSTD-RDIAN}
                                 & MTU-Net$_{23}$~\cite{IRSTD-MTU-Net}
                                 & ABC$_{23}$~\cite{IRSTD-ABC}
                                 & SeRankDet$_{24}$~\cite{IRSTD-SeRankDet}
                                 & MSHNet$_{24}$~\cite{IRSTD-MSHNet}
                                 & MRF3Net$_{24}$~\cite{IRSTD-MRF3Net}
                                 & SCTransNet$_{24}$~\cite{IRSTD-SCTransNet}
                                 & RPCANet$_{24}$~\cite{IRSTD-RPCANet}                                                                                                                                                                                                                                                                                     \\
\end{tabular}
}
\end{table*}

\parhead{Cross-Dataset Generalization.}
Cross-dataset analysis in Tab.~\ref{tab:metric_all_datasets} exposes critical generalization limitations.
Most models exhibit severe performance degradation when they are tested on other datasets beyond their training one.
Notably, IRSTD1k-trained models achieve better performance on SIRST$_{TE}$ than on IRSTD1k$_{TE}$.
And some IRSTD1k-trained models (\eg, MRF3Net with 0.694 hIoU) even surpass counterparts trained on SIRST$_{TR}$ (UIUNet: 0.679 hIoU; RPCANet: 0.675 hIoU) when evaluated on SIRST$_{TE}$.
This phenomenon may be attributed to IRSTD1k primarily focuses on sky-background scenarios from SIRST, while introducing additional challenging ground scenarios (\eg, woods and buildings).
These complexities make IRSTD1k$_{TE}$ inherently more difficult than SIRST$_{TE}$.
Current single-dataset evaluation frameworks fail to expose such critical performance limitations.
Establishing cross-dataset evaluation protocols would not only drive the community to prioritize model generalization across diverse scenarios, but also incentivize dataset creators to enhance scenario and target diversity.
This paradigm shift addresses the current oversight in robustness validation and fosters progress in domain adaptation research.

\begin{table*}[!t]
    \centering
    \caption{Cross-dataset error analysis for IRSTD1k$_{TR}$~\cite{IRSTD-ISNet}-trained models.
        See Tab.~\ref{tab:total_error_datasets} for details.}
    \label{tab:error_datasets}
    \resizebox{0.98\linewidth}{!}{\begin{tabular}{rr|rrrrrrrrrrrrrr}
    \toprule[2pt]
     &
     & ACM$_{21}$~\cite{IRSTD-ACM-nIoU}
     & FC3Net$_{22}$~\cite{IRSTD-FC3Net}
     & DNANet$_{22}$~\cite{IRSTD-DNANet}
     & ISNet$_{22}$~\cite{IRSTD-ISNet}
     & AGPCNet$_{23}$~\cite{IRSTD-AGPCNet}
     & UIUNet$_{23}$~\cite{IRSTD-UIUNet}
     & RDIAN$_{23}$~\cite{IRSTD-RDIAN}
     & MTU-Net$_{23}$~\cite{IRSTD-MTU-Net}
     & ABC$_{23}$~\cite{IRSTD-ABC}
     & SeRankDet$_{24}$~\cite{IRSTD-SeRankDet}
     & MSHNet$_{24}$~\cite{IRSTD-MSHNet}
     & MRF3Net$_{24}$~\cite{IRSTD-MRF3Net}
     & SCTransNet$_{24}$~\cite{IRSTD-SCTransNet}
     & RPCANet$_{24}$~\cite{IRSTD-RPCANet}                                                                                                                                                                               \\
    \midrule[1pt]
    \multirow{9}{*}{\rotatebox{90}{IRSTD1k$_{TE}$~\cite{IRSTD-ISNet}}}
     & IoU$^{seg}_{pix}\uparrow$                 & 5.387e-01 & 6.172e-01 & 6.694e-01 & 5.476e-01 & 6.139e-01 & 6.317e-01 & 6.529e-01 & 6.543e-01 & 6.361e-01 & 6.755e-01 & 6.497e-01 & 6.742e-01 & 6.610e-01 & 6.607e-01 \\
     & $\mathbf{E}^{seg}_{MRG}\downarrow$        & 2.132e-03 & 4.340e-04 & 5.610e-04 & 4.830e-04 & 4.340e-04 & 4.440e-04 & 5.410e-04 & 4.580e-04 & 4.680e-04 & 4.250e-04 & 0.000e+00 & 4.420e-04 & 4.770e-04 & 4.570e-04 \\
     & $\mathbf{E}^{seg}_{ITF}\downarrow$        & 2.630e-01 & 1.431e-01 & 1.669e-01 & 1.089e-01 & 1.330e-01 & 1.048e-01 & 1.470e-01 & 1.855e-01 & 1.017e-01 & 1.463e-01 & 1.186e-01 & 2.039e-01 & 1.135e-01 & 1.595e-01 \\
     & $\mathbf{E}^{seg}_{PCP}\downarrow$        & 1.962e-01 & 2.393e-01 & 1.631e-01 & 3.430e-01 & 2.527e-01 & 2.630e-01 & 1.996e-01 & 1.597e-01 & 2.618e-01 & 1.778e-01 & 2.317e-01 & 1.215e-01 & 2.250e-01 & 1.794e-01 \\
    \cmidrule{2-16}
     & IoU$^{loc}_{tgt}\uparrow$                 & 6.613e-01 & 6.205e-01 & 8.318e-01 & 8.088e-01 & 8.077e-01 & 8.390e-01 & 7.832e-01 & 7.541e-01 & 7.982e-01 & 7.694e-01 & 8.450e-01 & 8.207e-01 & 8.131e-01 & 7.112e-01 \\
     & $\mathbf{E}^{loc}_{S2M}\downarrow$        & 8.000e-03 & 2.387e-03 & 9.174e-03 & 2.941e-03 & 2.959e-03 & 3.096e-03 & 8.671e-03 & 5.405e-03 & 2.924e-03 & 2.778e-03 & 0.000e+00 & 6.079e-03 & 2.967e-03 & 2.674e-03 \\
     & $\mathbf{E}^{loc}_{M2S}\downarrow$        & 0.000e+00 & 2.387e-03 & 0.000e+00 & 0.000e+00 & 0.000e+00 & 3.096e-03 & 0.000e+00 & 0.000e+00 & 0.000e+00 & 0.000e+00 & 0.000e+00 & 6.079e-03 & 0.000e+00 & 0.000e+00 \\
     & $\mathbf{E}^{loc}_{ITF}\downarrow$        & 2.080e-01 & 2.888e-01 & 9.174e-02 & 1.265e-01 & 1.213e-01 & 7.740e-02 & 1.416e-01 & 1.973e-01 & 1.316e-01 & 1.750e-01 & 9.726e-02 & 9.119e-02 & 1.187e-01 & 2.059e-01 \\
     & $\mathbf{E}^{loc}_{PCP}\downarrow$        & 1.227e-01 & 8.592e-02 & 6.728e-02 & 6.177e-02 & 6.805e-02 & 7.740e-02 & 6.647e-02 & 4.324e-02 & 6.725e-02 & 5.278e-02 & 5.775e-02 & 7.599e-02 & 6.528e-02 & 8.021e-02 \\
    \midrule[0.5pt]
    \multirow{9}{*}{\rotatebox{90}{SIRST$_{TE}$~\cite{IRSTD-ACM-nIoU}}}
     & IoU$^{seg}_{pix}\uparrow$                 & 6.254e-01 & 7.317e-01 & 7.391e-01 & 7.113e-01 & 7.260e-01 & 6.857e-01 & 7.550e-01 & 7.564e-01 & 7.405e-01 & 7.480e-01 & 7.184e-01 & 7.658e-01 & 7.036e-01 & 7.575e-01 \\
     & $\mathbf{E}^{seg}_{MRG}\downarrow$        & 0.000e+00 & 0.000e+00 & 0.000e+00 & 0.000e+00 & 0.000e+00 & 6.040e-03 & 0.000e+00 & 0.000e+00 & 0.000e+00 & 0.000e+00 & 0.000e+00 & 0.000e+00 & 0.000e+00 & 0.000e+00 \\
     & $\mathbf{E}^{seg}_{ITF}\downarrow$        & 1.488e-01 & 7.411e-02 & 5.026e-02 & 7.203e-02 & 8.022e-02 & 4.956e-02 & 7.283e-02 & 9.772e-02 & 4.874e-02 & 6.509e-02 & 5.877e-02 & 8.256e-02 & 4.715e-02 & 6.688e-02 \\
     & $\mathbf{E}^{seg}_{PCP}\downarrow$        & 2.257e-01 & 1.941e-01 & 2.106e-01 & 2.167e-01 & 1.937e-01 & 2.587e-01 & 1.721e-01 & 1.459e-01 & 2.107e-01 & 1.869e-01 & 2.229e-01 & 1.516e-01 & 2.493e-01 & 1.756e-01 \\
    \cmidrule{2-16}
     & IoU$^{loc}_{tgt}\uparrow$                 & 6.689e-01 & 5.333e-01 & 9.292e-01 & 9.469e-01 & 9.391e-01 & 9.386e-01 & 8.537e-01 & 9.160e-01 & 9.068e-01 & 9.145e-01 & 8.667e-01 & 9.068e-01 & 8.468e-01 & 7.891e-01 \\
     & $\mathbf{E}^{loc}_{S2M}\downarrow$        & 0.000e+00 & 0.000e+00 & 0.000e+00 & 0.000e+00 & 0.000e+00 & 1.754e-02 & 0.000e+00 & 0.000e+00 & 0.000e+00 & 0.000e+00 & 0.000e+00 & 0.000e+00 & 0.000e+00 & 0.000e+00 \\
     & $\mathbf{E}^{loc}_{M2S}\downarrow$        & 6.623e-03 & 0.000e+00 & 0.000e+00 & 8.850e-03 & 0.000e+00 & 0.000e+00 & 0.000e+00 & 8.403e-03 & 0.000e+00 & 0.000e+00 & 0.000e+00 & 8.475e-03 & 0.000e+00 & 0.000e+00 \\
     & $\mathbf{E}^{loc}_{ITF}\downarrow$        & 2.715e-01 & 3.944e-01 & 3.540e-02 & 2.655e-02 & 5.217e-02 & 4.386e-02 & 1.138e-01 & 7.563e-02 & 7.627e-02 & 6.838e-02 & 9.167e-02 & 6.780e-02 & 1.210e-01 & 1.484e-01 \\
     & $\mathbf{E}^{loc}_{PCP}\downarrow$        & 5.298e-02 & 7.222e-02 & 3.540e-02 & 1.770e-02 & 8.696e-03 & 0.000e+00 & 3.252e-02 & 0.000e+00 & 1.695e-02 & 1.709e-02 & 4.167e-02 & 1.695e-02 & 3.226e-02 & 6.250e-02 \\
    \midrule[0.5pt]
    \multirow{9}{*}{\rotatebox{90}{NUDT$_{TE}$~\cite{IRSTD-DNANet}}}
     & IoU$^{seg}_{pix}\uparrow$                 & 6.170e-01 & 6.490e-01 & 7.250e-01 & 7.177e-01 & 6.672e-01 & 6.491e-01 & 6.475e-01 & 6.569e-01 & 7.145e-01 & 7.075e-01 & 6.670e-01 & 7.032e-01 & 6.504e-01 & 6.966e-01 \\
     & $\mathbf{E}^{seg}_{MRG}\downarrow$        & 1.087e-03 & 1.249e-03 & 0.000e+00 & 1.242e-03 & 1.124e-03 & 1.110e-03 & 1.270e-03 & 1.291e-03 & 0.000e+00 & 1.248e-03 & 0.000e+00 & 1.284e-03 & 1.176e-03 & 1.456e-03 \\
     & $\mathbf{E}^{seg}_{ITF}\downarrow$        & 1.875e-01 & 1.287e-01 & 1.127e-01 & 1.463e-01 & 1.261e-01 & 9.163e-02 & 8.371e-02 & 1.352e-01 & 1.242e-01 & 1.275e-01 & 9.738e-02 & 1.079e-01 & 7.780e-02 & 1.093e-01 \\
     & $\mathbf{E}^{seg}_{PCP}\downarrow$        & 1.944e-01 & 2.210e-01 & 1.622e-01 & 1.347e-01 & 2.056e-01 & 2.581e-01 & 2.676e-01 & 2.066e-01 & 1.613e-01 & 1.637e-01 & 2.357e-01 & 1.876e-01 & 2.706e-01 & 1.927e-01 \\
    \cmidrule{2-16}
     & IoU$^{loc}_{tgt}\uparrow$                 & 4.443e-01 & 5.129e-01 & 7.261e-01 & 6.042e-01 & 6.829e-01 & 7.040e-01 & 6.301e-01 & 5.569e-01 & 6.767e-01 & 6.580e-01 & 6.673e-01 & 6.884e-01 & 6.045e-01 & 4.944e-01 \\
     & $\mathbf{E}^{loc}_{S2M}\downarrow$        & 1.311e-03 & 1.522e-03 & 0.000e+00 & 1.751e-03 & 1.876e-03 & 1.898e-03 & 1.689e-03 & 1.517e-03 & 0.000e+00 & 1.848e-03 & 0.000e+00 & 1.866e-03 & 1.727e-03 & 1.600e-03 \\
     & $\mathbf{E}^{loc}_{M2S}\downarrow$        & 1.180e-02 & 1.218e-02 & 1.149e-02 & 2.627e-02 & 4.878e-02 & 2.656e-02 & 7.264e-02 & 7.436e-02 & 2.256e-02 & 1.294e-02 & 5.176e-02 & 5.597e-02 & 7.081e-02 & 2.080e-02 \\
     & $\mathbf{E}^{loc}_{ITF}\downarrow$        & 4.273e-01 & 3.364e-01 & 1.686e-01 & 2.242e-01 & 1.482e-01 & 1.613e-01 & 2.044e-01 & 2.762e-01 & 1.729e-01 & 1.959e-01 & 1.571e-01 & 1.455e-01 & 1.900e-01 & 2.944e-01 \\
     & $\mathbf{E}^{loc}_{PCP}\downarrow$        & 1.153e-01 & 1.370e-01 & 9.387e-02 & 1.436e-01 & 1.182e-01 & 1.063e-01 & 9.122e-02 & 9.105e-02 & 1.278e-01 & 1.312e-01 & 1.238e-01 & 1.082e-01 & 1.330e-01 & 1.888e-01 \\
    \bottomrule[2pt]
\end{tabular}
}
\end{table*}

\begin{figure*}[!t]
    \centering
    \subfloat[Localization errors.]{\includegraphics[width=0.49\linewidth]{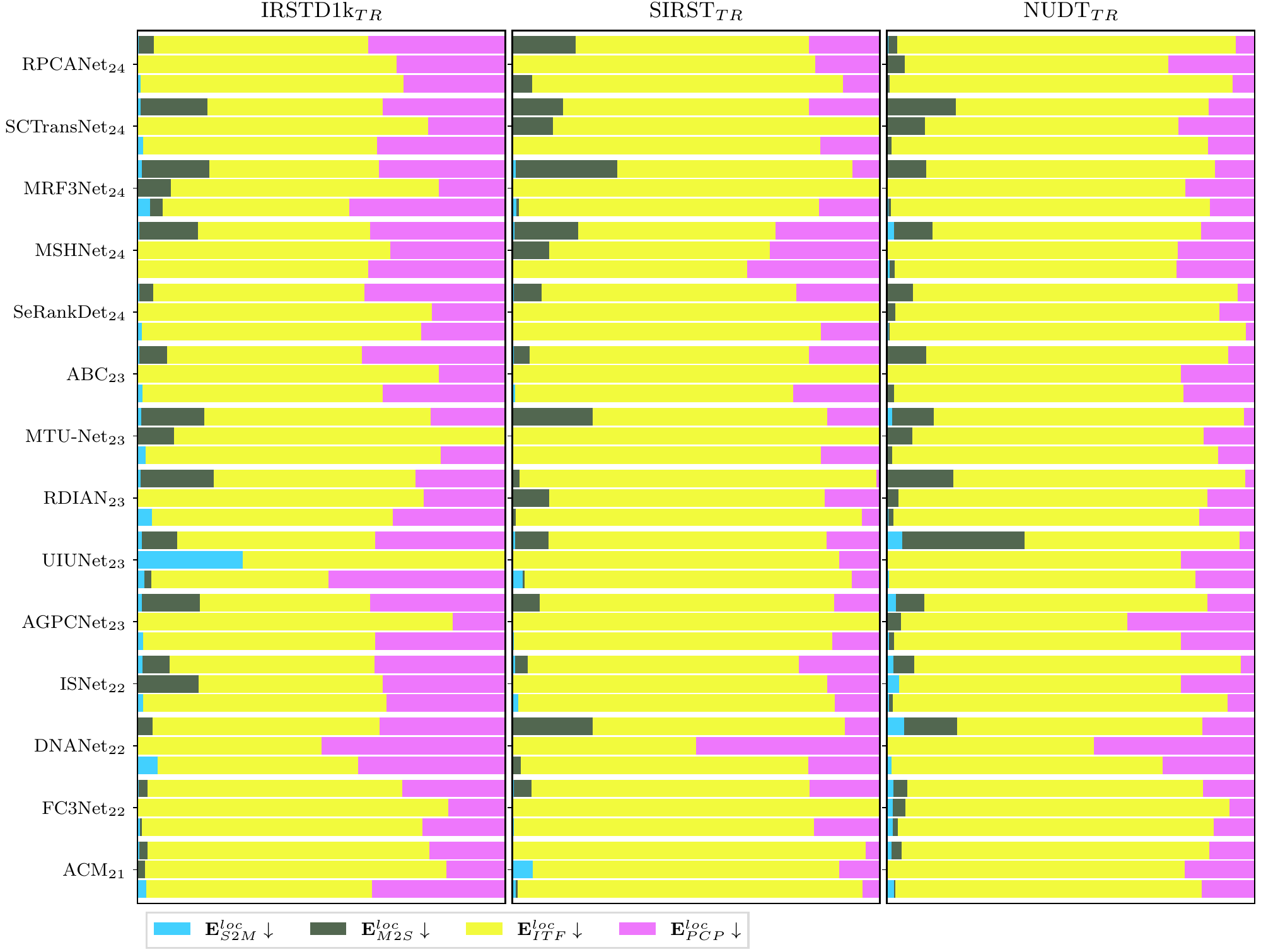}}
    \subfloat[Segmentation errors.]{\includegraphics[width=0.49\linewidth]{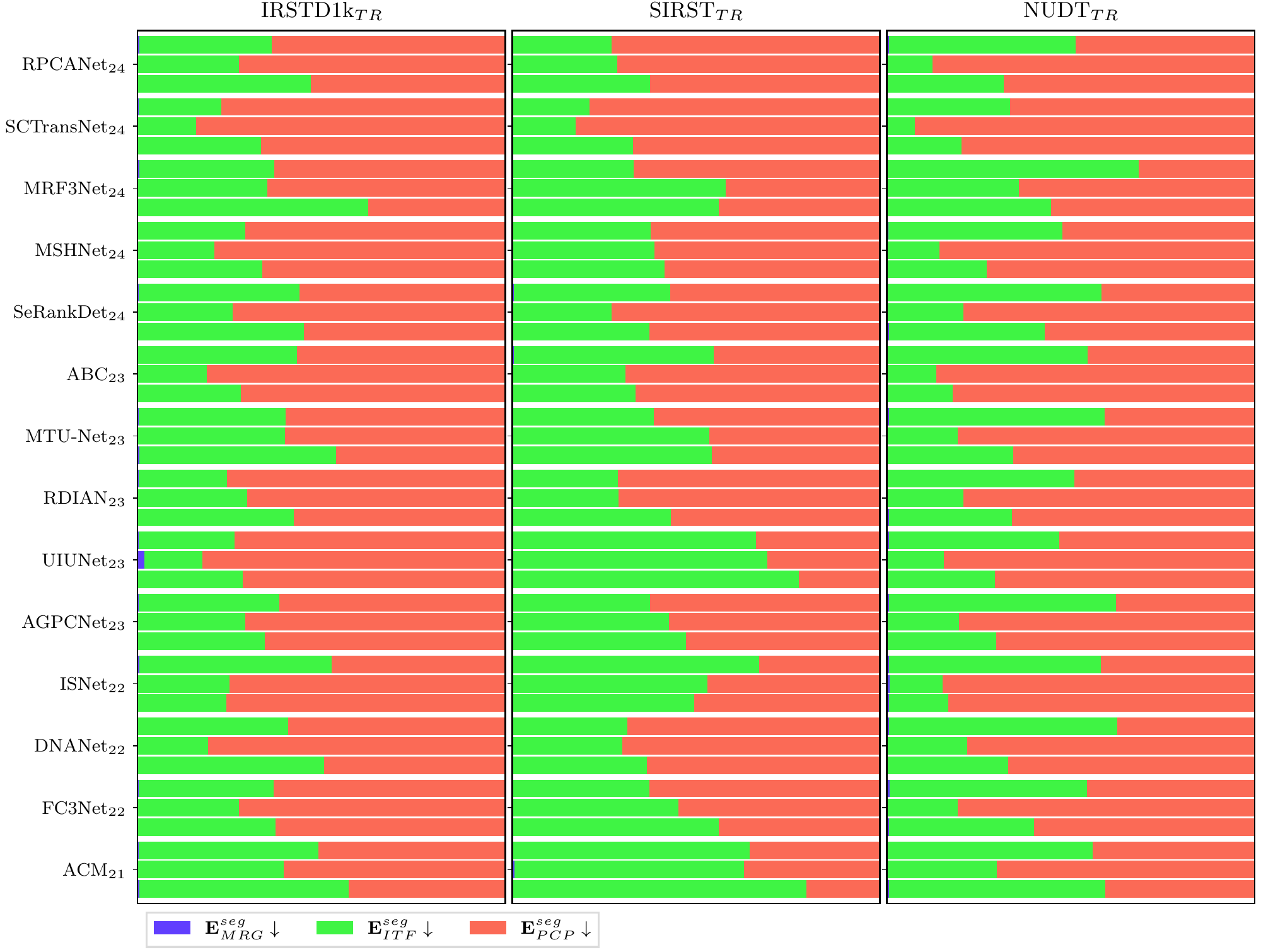}}
    \caption{Cross-dataset error ratios corresponding to IRSTD1k$_{TE}$, SIRST$_{TE}$, and NUDT$_{TE}$ from bottom to top for the models trained on different datasets.}
    \label{fig:error_ratio}
    \vspace{-2ex}
\end{figure*}

\parhead{Dense Multi-Object Scenarios.}
As shown in Fig.~\ref{fig:error_ratio}, these errors are predominantly dominated by interference- and perception-related terms.
More specifically, interference item dominates the localization error, while perception item dominates the segmentation error.
The former reflects the limited capability of current algorithms in suppressing background interference.
The latter is closely tied to the intrinsic challenges of infrared small targets, such as blurred boundaries and complex structure patterns, which may lead to insufficient holistic perception during prediction.
Furthermore, the analysis reveals that matching errors (\ie, $\mathbf{E}^{loc}_{S2M}$ and $\mathbf{E}^{loc}_{M2S}$) and prediction merging errors (\ie, $\mathbf{E}^{seg}_{MRG}$), often arising in dense multi-target scenarios or cases with complex target structures, exhibit relatively low proportions due to current data limitations.
However, experiments in Sec.~\ref{supp:synthetic_data} using synthetic data reveal that they still make up a significant portion of total error.
Their impact cannot be overlooked, as they represent critical failure modes in real-world applications.
For instance, single-to-multi or multi-to-single errors may disrupt downstream tasks like counting, and merged segmentation masks could degrade the reliability of target region decisions.
This highlights the necessity of future research to address such edge cases while balancing dataset biases.

\section{Conclusion}

Our work carefully reviews the limitations of existing evaluation protocols in IRSTD.
And to this end, we introduce a novel hybrid-level performance metric and a systematic error analysis method, emphasizing the necessity of cross-dataset validation in evaluation.
The proposed metric holistically models hybrid localization-segmentation information to comprehensively characterize the algorithm capabilities.
Coupled with this, our error analysis provides fine-grained diagnostic insights into failure modes, thereby deepening the understanding of model performance.
Through extensive within- and cross-dataset experiments on existing benchmarks, we demonstrate the effectiveness and interpretative power of our framework in evaluating the performance of IRSTD methods.
By establishing an enhanced analytical paradigm, our work aims to advance the methodological foundation of IRSTD algorithm development and evaluation, with important implications for guiding more robust model design and validation in practical applications.

    {
        \small
        \bibliographystyle{plain}
        \bibliography{abbr,main}

\begin{thebibliography}{10}

\bibitem{IRSTD-TraditionalSuppression-NWMTH}
Xiangzhi Bai and Fugen Zhou.
\newblock Analysis of new top-hat transformation and the application for infrared dim small target detection.
\newblock {\em Pattern Recognition}, 43(6):2145--2156, 2010.

\bibitem{STD-SatelliteImaging}
Michael Bruno, Alexander Sutin, Kil~Woo Chung, Alexander Sedunov, Nikolay Sedunov, Hady Salloum, Hans Graber, and Paul Mallas.
\newblock Satellite imaging and passive acoustics in layered approach for small boat detection and classification.
\newblock {\em Marine Technology Society Journal}, 45(3):77--87, 2011.

\bibitem{IRSTD-OceanEnvironment}
Zhaoyang Cao, Xuan Kong, Qiang Zhu, Siying Cao, and Zhenming Peng.
\newblock Infrared dim target detection via mode-k1k2 extension tensor tubal rank under complex ocean environment.
\newblock {\em ISPRS Journal of Photogrammetry and Remote Sensing}, 181:167--190, 2021.

\bibitem{IRSTDSurvey}
Yongbo Cheng, Xuefeng Lai, Yucheng Xia, and Jinmei Zhou.
\newblock Infrared dim small target detection networks: A review.
\newblock {\em Sensors}, 24(12):3885, June 2024.

\bibitem{linear_sum_assignment}
David~F. Crouse.
\newblock On implementing 2d rectangular assignment algorithms.
\newblock {\em {IEEE} Transactions on Aerospace and Electronic Systems}, 52(4):1679--1696, August 2016.

\bibitem{IRSTD-SeRankDet}
Yimian Dai, Peiwen Pan, Yulei Qian, Yuxuan Li, Xiang Li, Jian Yang, and Huan Wang.
\newblock Pick of the bunch: Detecting infrared small targets beyond hit-miss trade-offs via selective rank-aware attention.
\newblock {\em {IEEE} Transactions on Geoscience and Remote Sensing}, 62:1--15, 2024.

\bibitem{IRSTD-TraditionalDecomposition-RIPT}
Yimian Dai and Yiquan Wu.
\newblock Reweighted infrared patch-tensor model with both nonlocal and local priors for single-frame small target detection.
\newblock {\em {IEEE} Journal of Selected Topics in Applied Earth Observations and Remote Sensing}, 10(8):3752--3767, 2017.

\bibitem{IRSTD-TraditionalDecomposition-NIPPS}
Yimian Dai, Yiquan Wu, Yu~Song, and Jun Guo.
\newblock Non-negative infrared patch-image model: Robust target-background separation via partial sum minimization of singular values.
\newblock {\em Infrared Physics \& Technology}, 81:182--194, 2017.

\bibitem{IRSTD-ACM-nIoU}
Yimian Dai, Yiquan Wu, Fei Zhou, and Kobus Barnard.
\newblock Asymmetric contextual modulation for infrared small target detection.
\newblock In {\em Proceedings of the IEEE Winter Conference on Applications of Computer Vision}, pages 949--958, Jan 2021.

\bibitem{IRSTD-TraditionalSuppression-WLDM}
He~Deng, Xianping Sun, Maili Liu, Chaohui Ye, and Xin Zhou.
\newblock Small infrared target detection based on weighted local difference measure.
\newblock {\em {IEEE} Transactions on Geoscience and Remote Sensing}, 54(7):4204--4214, 2016.

\bibitem{PraNet}
Deng-Ping Fan, Ge-Peng Ji, Tao Zhou, Geng Chen, Huazhu Fu, Jianbing Shen, and Ling Shao.
\newblock Pranet: Parallel reverse attention network for polyp segmentation.
\newblock In {\em International Conference on Medical Image Computing and Computer-Assisted Intervention}, 2020.

\bibitem{IRSTD-TraditionalDecomposition-IPI}
Chenqiang Gao, Deyu Meng, Yi~Yang, Yongtao Wang, Xiaofang Zhou, and Alexander~G Hauptmann.
\newblock Infrared patch-image model for small target detection in a single image.
\newblock {\em {IEEE} Transactions on Image Processing}, 22(12):4996--5009, 2013.

\bibitem{IRSTD-TraditionalSuppression-RLCM}
Jinhui Han, Kun Liang, Bo~Zhou, Xinying Zhu, Jie Zhao, and Linlin Zhao.
\newblock Infrared small target detection utilizing the multiscale relative local contrast measure.
\newblock {\em {IEEE} Geoscience and Remote Sensing Letters}, 15(4):612--616, 2018.

\bibitem{IRSTD-TraditionalSuppression-TLLCM}
Jinhui Han, Sibang Liu, Gang Qin, Qian Zhao, Honghui Zhang, and Nana Li.
\newblock A local contrast method combined with adaptive background estimation for infrared small target detection.
\newblock {\em {IEEE} Geoscience and Remote Sensing Letters}, 16(9):1442--1446, 2019.

\bibitem{IRSTD-TraditionalSuppression-ILCM}
Jinhui Han, Yong Ma, Bo~Zhou, Fan Fan, Kun Liang, and Yu~Fang.
\newblock A robust infrared small target detection algorithm based on human visual system.
\newblock {\em {IEEE} Geoscience and Remote Sensing Letters}, 11(12):2168--2172, 2014.

\bibitem{IRSTDSurvey-SegmentationNetworks}
Renke Kou, Chunping Wang, Zhenming Peng, Zhihe Zhao, Yaohong Chen, Jinhui Han, Fuyu Huang, Ying Yu, and Qiang Fu.
\newblock Infrared small target segmentation networks: A survey.
\newblock {\em Pattern Recognition}, 143:109788, November 2023.

\bibitem{IRSTD-DNANet}
Boyang Li, Chao Xiao, Longguang Wang, Yingqian Wang, Zaiping Lin, Miao Li, Wei An, and Yulan Guo.
\newblock Dense nested attention network for infrared small target detection.
\newblock {\em {IEEE} Transactions on Image Processing}, 32:1745--1758, 2023.

\bibitem{MIRSTD-DTUM}
Ruojing Li, Wei An, Chao Xiao, Boyang Li, Yingqian Wang, Miao Li, and Yulan Guo.
\newblock Direction-coded temporal u-shape module for multiframe infrared small target detection.
\newblock {\em {IEEE} Transactions on Neural Networks and Learning Systems}, 36(1):555--568, 2025.

\bibitem{MIRSTD-DeepPro}
Ruojing Li, Wei An, Xinyi Ying, Yingqian Wang, Yimian Dai, Longguang Wang, Miao Li, Yulan Guo, and Li~Liu.
\newblock Probing deep into temporal profile makes the infrared small target detector much better.
\newblock {\em CoRR}, abs/2506.12766, 2025.

\bibitem{IRSTD-MSHNet}
Qiankun Liu, Rui Liu, Bolun Zheng, Hongkui Wang, and Ying FU.
\newblock Infrared small target detection with scale and location sensitivity.
\newblock In {\em Proceedings of the IEEE Conference on Computer Vision and Pattern Recognition}, pages 17490--17499, June 2024.

\bibitem{FCN}
Jonathan Long, Evan Shelhamer, and Trevor Darrell.
\newblock Fully convolutional networks for semantic segmentation.
\newblock In {\em Proceedings of the IEEE Conference on Computer Vision and Pattern Recognition}, 2015.

\bibitem{IRSTD-ABC}
Peiwen Pan, Huan Wang, Chenyi Wang, and Chang Nie.
\newblock Abc: Attention with bilinear correlation for infrared small target detection.
\newblock In {\em Proceedings of the IEEE International Conference on Multimedia and Expo}, pages 2381--2386, July 2023.

\bibitem{IRSTD-TraditionalSuppression-FKRW}
Yao Qin, Lorenzo Bruzzone, Chengqiang Gao, and Biao Li.
\newblock Infrared small target detection based on facet kernel and random walker.
\newblock {\em {IEEE} Transactions on Geoscience and Remote Sensing}, 57(9):7104--7118, 2019.

\bibitem{IRSTD-TraditionalSuppression-GSWLCM}
Zhaobing Qiu, Yong Ma, Fan Fan, Jun Huang, and Lang Wu.
\newblock Global sparsity-weighted local contrast measure for infrared small target detection.
\newblock {\em {IEEE} Geoscience and Remote Sensing Letters}, 19:1--5, 2022.

\bibitem{Unet}
Olaf Ronneberger, Philipp Fischer, and Thomas Brox.
\newblock U-net: Convolutional networks for biomedical image segmentation.
\newblock In {\em International Conference on Medical Image Computing and Computer-Assisted Intervention}, 2015.

\bibitem{IRSTD-RDIAN}
Heng Sun, Junxiang Bai, Fan Yang, and Xiangzhi Bai.
\newblock Receptive-field and direction induced attention network for infrared dim small target detection with a large-scale dataset irdst.
\newblock {\em {IEEE} Transactions on Geoscience and Remote Sensing}, 61:1--13, 2023.

\bibitem{IRSTD-Maritime}
Fan Wang, Weixian Qian, Ye~Qian, Chao Ma, He~Zhang, Jiajie Wang, Minjie Wan, and Kan Ren.
\newblock Maritime infrared small target detection based on the appearance stable isotropy measure in heavy sea clutter environments.
\newblock {\em Sensors}, 23(24):9838, 2023.

\bibitem{IRSTD-TraditionalSuppression-MPCM}
Yantao Wei, Xinge You, and Hong Li.
\newblock Multiscale patch-based contrast measure for small infrared target detection.
\newblock {\em Pattern Recognition}, 58:216--226, 2016.

\bibitem{IRSTD-RPCANet}
Fengyi Wu, Tianfang Zhang, Lei Li, Yian Huang, and Zhenming Peng.
\newblock Rpcanet: Deep unfolding rpca based infrared small target detection.
\newblock In {\em Proceedings of the IEEE Winter Conference on Applications of Computer Vision}, pages 4797--4806, Jan 2024.

\bibitem{IRSTD-MTU-Net}
Tianhao Wu, Boyang Li, Yihang Luo, Yingqian Wang, Chao Xiao, Ting Liu, Jungang Yang, Wei An, and Yulan Guo.
\newblock Mtu-net: Multilevel transunet for space-based infrared tiny ship detection.
\newblock {\em {IEEE} Transactions on Geoscience and Remote Sensing}, 61:1--15, 2023.

\bibitem{IRSTD-UIUNet}
Xin Wu, Danfeng Hong, and Jocelyn Chanussot.
\newblock Uiu-net: U-net in u-net for infrared small object detection.
\newblock {\em {IEEE} Transactions on Image Processing}, 32:364--376, 2023.

\bibitem{MIRSTD-RFR}
Xinyi Ying, Li~Liu, Zaiping Lin, Yangsi Shi, Yingqian Wang, Ruojing Li, Xu~Cao, Boyang Li, Shilin Zhou, and Wei An.
\newblock Infrared small target detection in satellite videos: {A} new dataset and a novel recurrent feature refinement framework.
\newblock {\em {IEEE} Transactions on Geoscience and Remote Sensing}, 63:1--18, 2025.

\bibitem{IRSTR-MoCoPnet}
Xinyi Ying, Yingqian Wang, Longguang Wang, Weidong Sheng, Li~Liu, Zaiping Lin, and Shilin Zhou.
\newblock Mocopnet: Exploring local motion and contrast priors for infrared small target super-resolution.
\newblock {\em CoRR}, abs/2201.01014, 2022.

\bibitem{IRSTD-SCTransNet}
Shuai Yuan, Hanlin Qin, Xiang Yan, Naveed Akhtar, and Ajmal Mian.
\newblock Sctransnet: Spatial-channel cross transformer network for infrared small target detection.
\newblock {\em {IEEE} Transactions on Geoscience and Remote Sensing}, 62:1--15, 2024.

\bibitem{IRSTD-TraditionalDecomposition-NRAM}
Landan Zhang, Lingbing Peng, Tianfang Zhang, Siying Cao, and Zhenming Peng.
\newblock Infrared small target detection via non-convex rank approximation minimization joint $\ell_{2,1}$ norm.
\newblock {\em Remote Sensing}, 10(11):1821, 2018.

\bibitem{IRSTD-TraditionalDecomposition-PSTNN}
Landan Zhang and Zhenming Peng.
\newblock Infrared small target detection based on partial sum of the tensor nuclear norm.
\newblock {\em Remote Sensing}, 11(4):382, 2019.

\bibitem{IRSTD-FC3Net}
Mingjin Zhang, Ke~Yue, Jing Zhang, Yunsong Li, and Xinbo Gao.
\newblock Exploring feature compensation and cross-level correlation for infrared small target detection.
\newblock In {\em Proceedings of the ACM International Conference on Multimedia}, 2022.

\bibitem{IRSTD-ISNet}
Mingjin Zhang, Rui Zhang, Yuxiang Yang, Haichen Bai, Jing Zhang, and Jie Guo.
\newblock Isnet: Shape matters for infrared small target detection.
\newblock In {\em Proceedings of the IEEE Conference on Computer Vision and Pattern Recognition}, pages 867--876, June 2022.

\bibitem{IRSTD-AGPCNet}
Tianfang Zhang, Lei Li, Siying Cao, Tian Pu, and Zhenming Peng.
\newblock Attention-guided pyramid context networks for detecting infrared small target under complex background.
\newblock {\em {IEEE} Transactions on Aerospace and Electronic Systems}, 59(4):4250--4261, Aug 2023.

\bibitem{IRSTD-TraditionalDecomposition-NOLC}
Tianfang Zhang, Hao Wu, Yuhan Liu, Lingbing Peng, Chunping Yang, and Zhenming Peng.
\newblock Infrared small target detection based on non-convex optimization with $l_p$-norm constraint.
\newblock {\em Remote Sensing}, 11(5):559, 2019.

\bibitem{IRSTD-MRF3Net}
Xiaohan Zhang, Xue Zhang, Si-Yuan Cao, Beinan Yu, Chenghao Zhang, and Hui-Liang Shen.
\newblock Mrf3net: An infrared small target detection network using multireceptive field perception and effective feature fusion.
\newblock {\em {IEEE} Transactions on Geoscience and Remote Sensing}, 62:1--14, 2024.

\end{thebibliography}
    }

\appendix

\section*{\LARGE Technical Appendices and Supplementary Material}

\section{OPDC Matching Strategy}

Current target-level IRSTD metrics suffer from overly strict distance-only filtering~\cite{IRSTD-DNANet}, where centroid matching frequently misjudges offset, fragmented, or connected predictions as shown in Fig.~\ref{fig:metric_comparison_1} and Fig.~\ref{fig:metric_comparison_2}.
By introducing overlap-priority constraint to enhance the matching mechanism, we propose the ``Overlap Priority with Distance Compensation'' (OPDC) strategy (Alg.~\ref{alg:matching}), which can effectively alleviates these limitations.

\begin{algorithm}[!h]
    \footnotesize
    \caption{OPDC Matching Strategy}
    \label{alg:matching}

    \KwIn{
    $\{T_G^m\}_{m}^{M}$: the mask set of $M$ targets extracted from GT map $G$;
    $\{T_P^n\}_{n}^{N}$: the mask set of $N$ targets extracted from prediction map $P$;
    $\text{MAX}$: a extremely large value used to avoid the algorithm choosing irrational matches;
    }
    \KwOut{
        $S_{TP}$: the matched index pair set;
        $S_{FN}$: the unmatched GT target index set;
        $S_{FP}$: the unmatched predicted target index set;
    }

    \tcp{1. Overlap Priority Constraint}
    Valid indicator $\mathbb{I} \in \mathbb{R}^{M \times N}, \mathbb{I}_{m,n} \in \{0,1\}, \mathbb{I}_{m,n} = 0, \forall m,n$\;
    Distance matrix $D \in \mathbb{R}^{M \times N}, D_{m,n} = \text{MAX}, \forall m,n$\;
    \For{$m=1$ \KwTo $M$}{
    \For{$n=1$ \KwTo $N$}{
    $D_{m,n} = \texttt{EuclideanDistance}(T_G^m, T_P^n)$\;

    \lIf{$|T_{G}^m \cap T_{P}^n| / |T_{G}^m \cup T_{P}^n| \ge 0.5$}{$\mathbb{I}_{m,n} = 1$}
    }
    }
    Initial match $A = \texttt{Assignment}(D)$\tcp*[r]{\texttt{scipy.optimize.linear\_sum\_assignment}~\cite{linear_sum_assignment}}
    $S_{TP}, S_{FN}, S_{FP} \gets A \cap \mathbb{I}$\tcp*[r]{Consider only pairwise relations that satisfy constraints.}

    \tcp{2. Distance-based Compensation}
    Valid indicator $\hat{\mathbb{I}} \in \mathbb{R}^{|S_{FN}| \times |S_{FP}|}, \hat{\mathbb{I}}_{m,n} \in \{0,1\}, \hat{\mathbb{I}}_{m,n} = 0, \forall m,n$\;
    Distance matrix $\hat{D} \in \mathbb{R}^{|S_{FN}| \times |S_{FP}|}, \hat{D}_{m,n} = \text{MAX}, \forall m,n$\;
    \For{$m=1$ \KwTo $|S_{FN}|$}{
    \For{$n=1$ \KwTo $|S_{FP}|$}{
    \If(\tcp*[f]{Following the setting in~\cite{IRSTD-DNANet}.})
    {$D_{S_{FN}^m, S_{FN}^n} < 3$}{
        $\hat{D}_{m,n} = D_{S_{FN}^m, S_{FN}^n}$\;
        $\hat{\mathbb{I}}_{m,n} = 1$\;
    }
    }
    }
    Compensation match $\hat{A} = \texttt{Assignment}(\hat{D})$\tcp*[r]{\texttt{scipy.optimize.linear\_sum\_assignment}~\cite{linear_sum_assignment}}

    $\hat{S}_{TP}, \hat{S}_{FN}, \hat{S}_{FP} \gets \hat{A} \cap \hat{\mathbb{I}}$\;

    $S_{TP} = S_{TP} \cup \hat{S}_{TP}$\tcp*[r]{Construct final matched index pair set.}
    $S_{FN} = S_{FN} \setminus \hat{S}_{TP}$\tcp*[r]{Construct final unmatched GT target index set.}
    $S_{FP} = S_{FP} \setminus \hat{S}_{TP}$\tcp*[r]{Construct final unmatched predicted target index set.}
\end{algorithm}

\section{Clearer Comparison}
\label{supp:clearer_comparison}

\begin{figure*}[!h]
    \centering
    \subfloat[]{\centering
        \label{fig:total_metric_comparison_1}
        \includegraphics[width=0.32\linewidth]{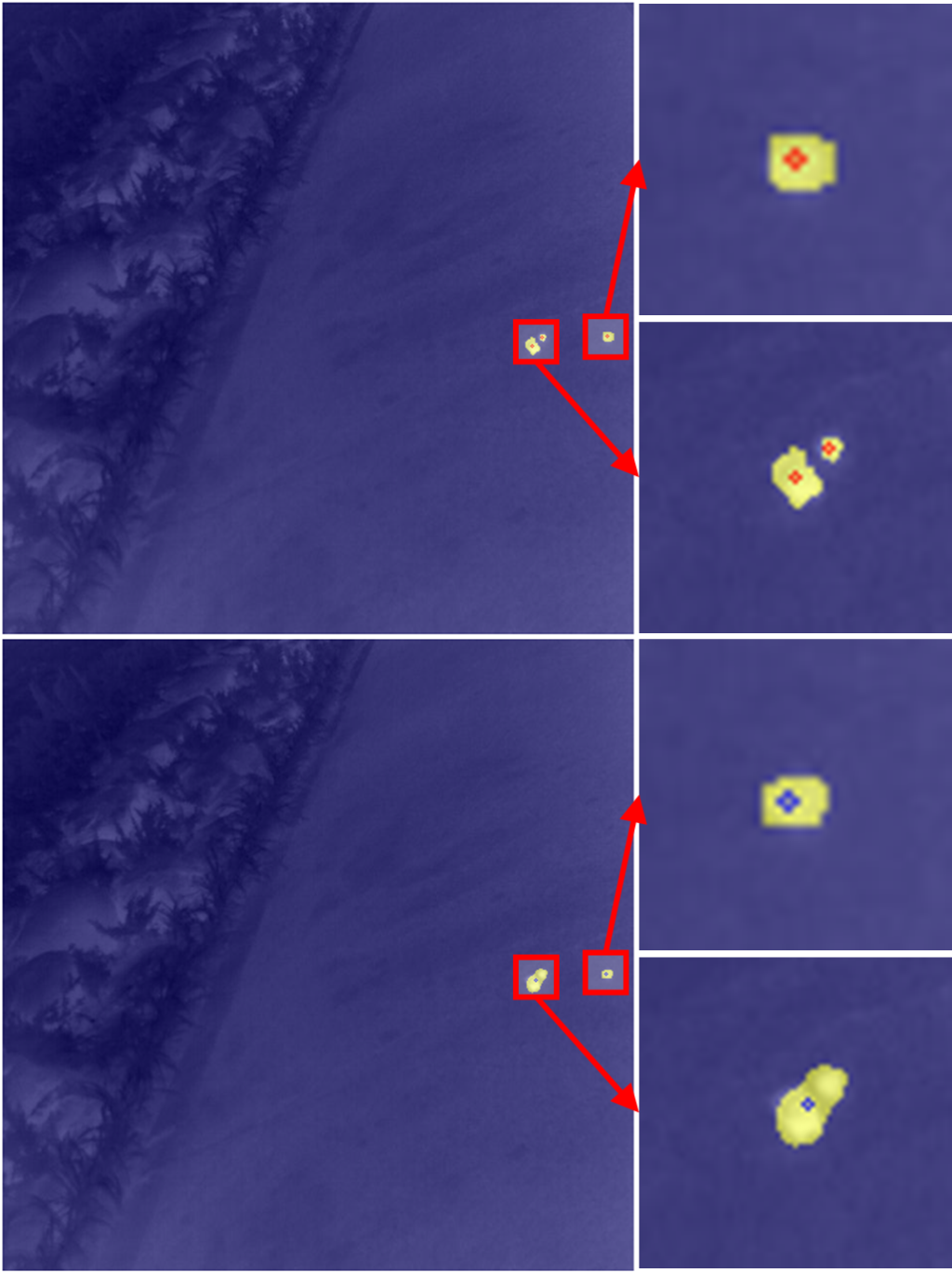}}
    \subfloat[]{\centering
        \label{fig:total_metric_comparison_2}
        \includegraphics[width=0.32\linewidth]{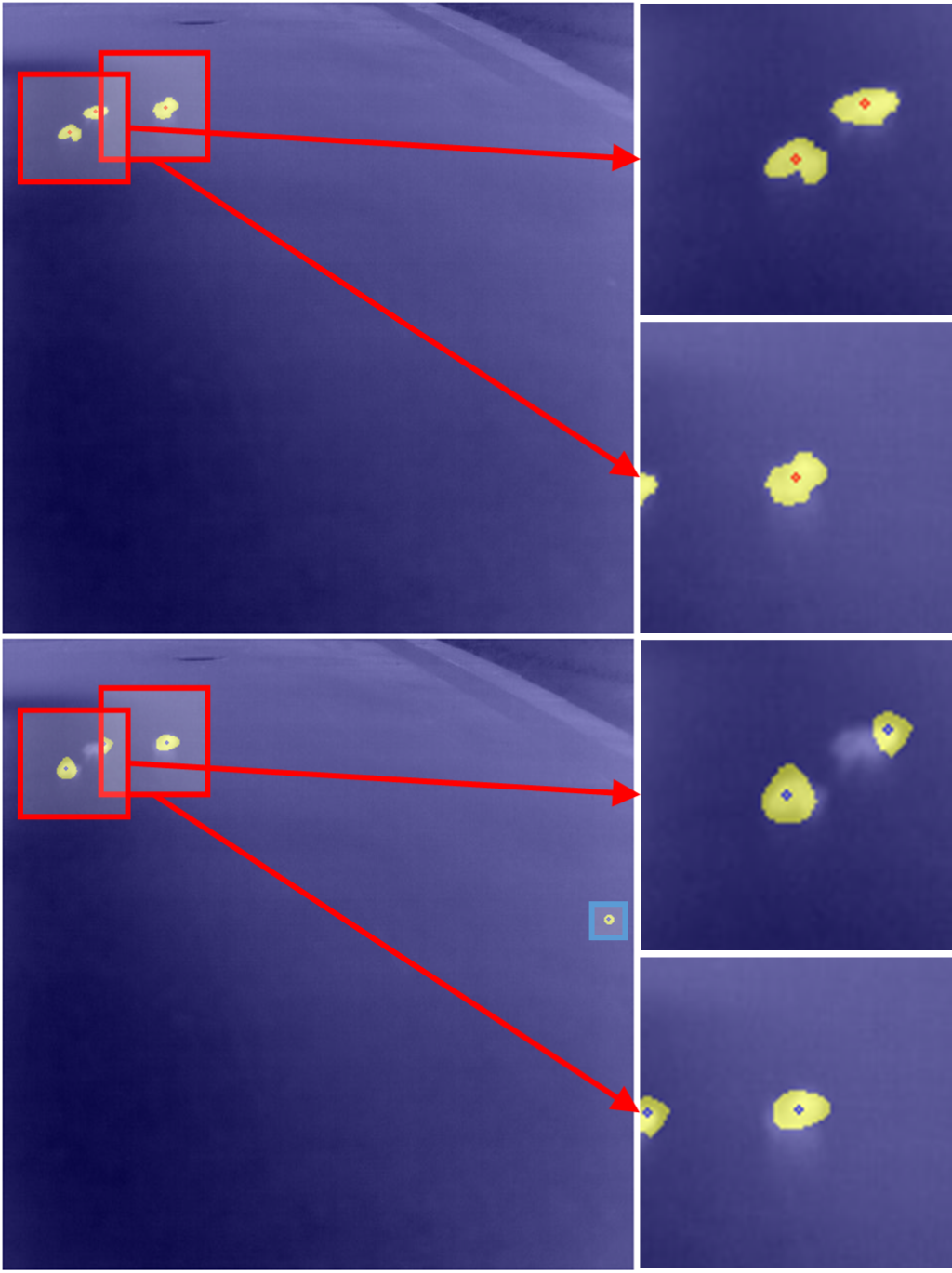}}
    \subfloat[]{\centering
        \label{fig:total_metric_comparison_3}
        \includegraphics[width=0.32\linewidth]{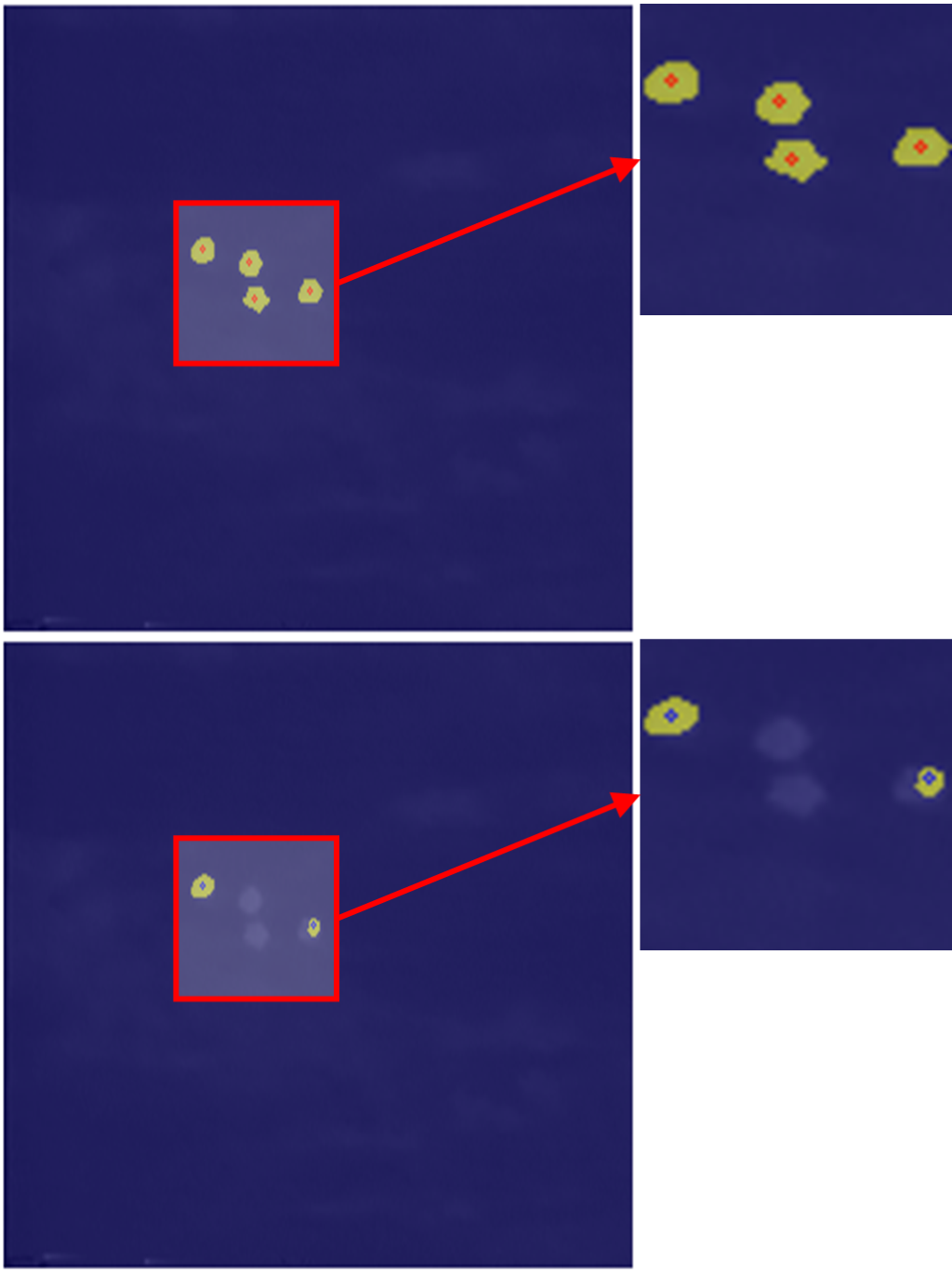}}
    \par
    \subfloat[]{\centering
        \label{fig:total_metric_comparison_4}
        \includegraphics[width=0.32\linewidth]{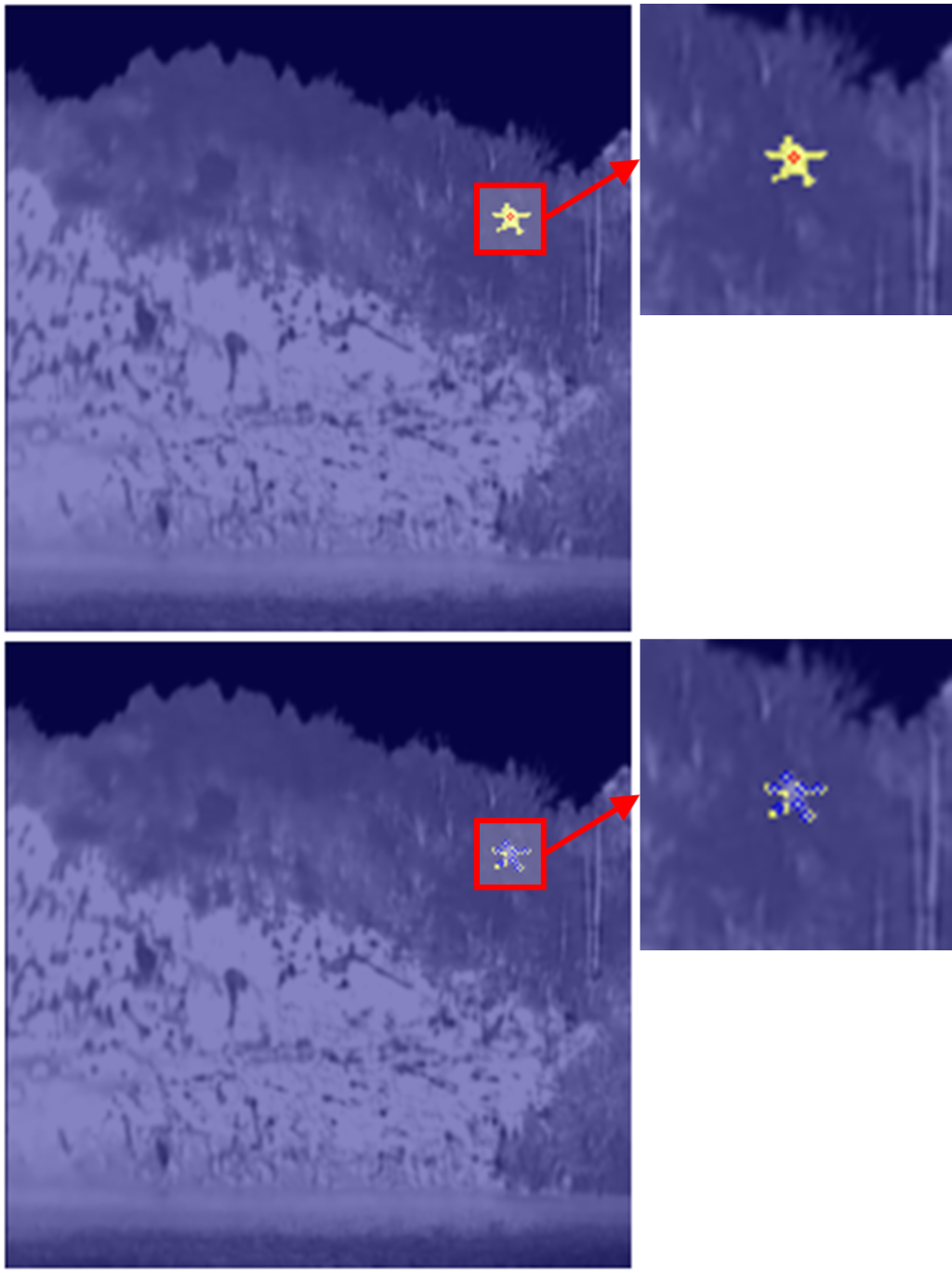}}
    \subfloat[]{\centering
        \label{fig:total_metric_comparison_5}
        \includegraphics[width=0.32\linewidth]{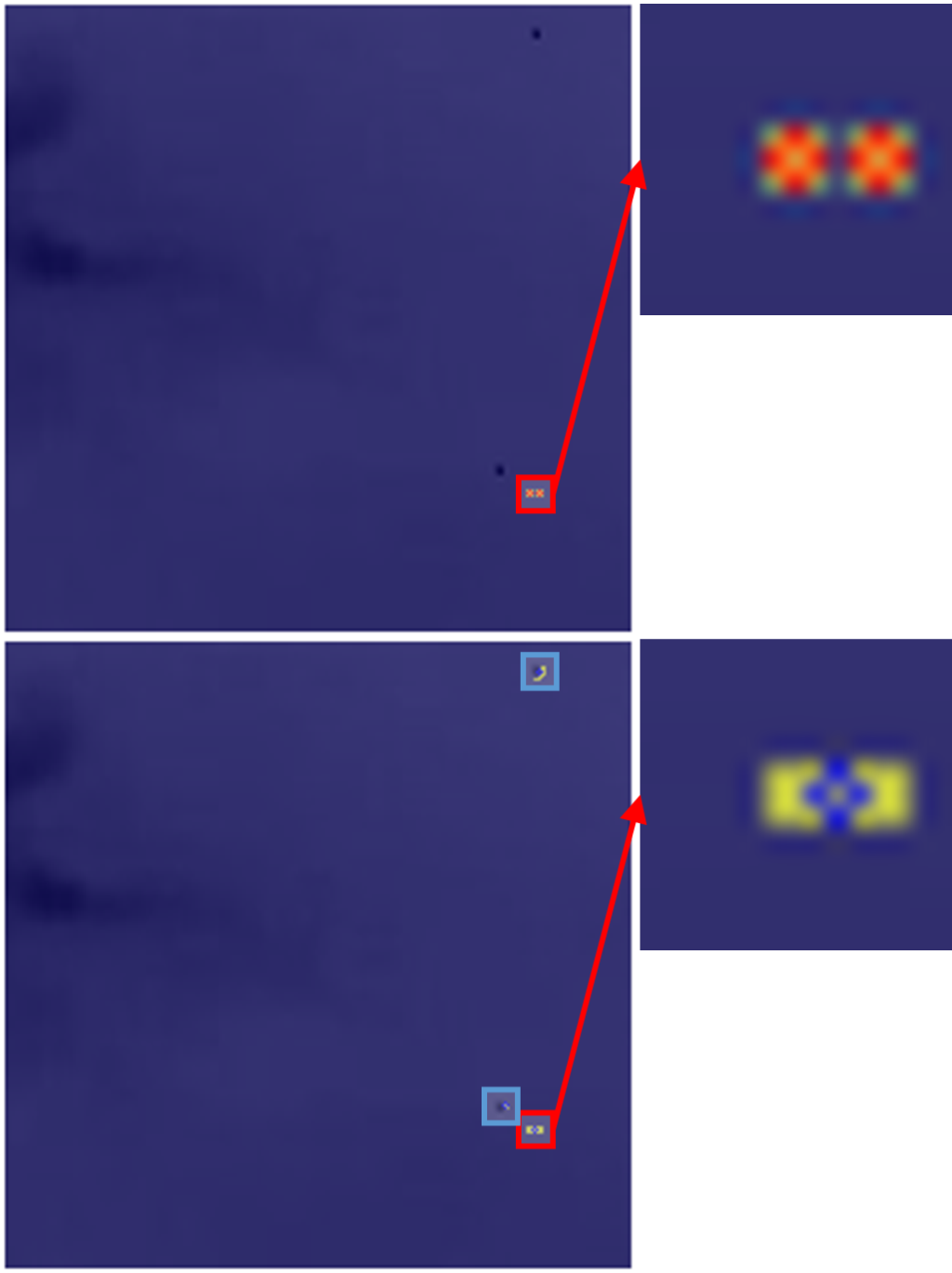}}
    \subfloat[]{\centering
        \label{fig:total_metric_comparison_6}
        \includegraphics[width=0.32\linewidth]{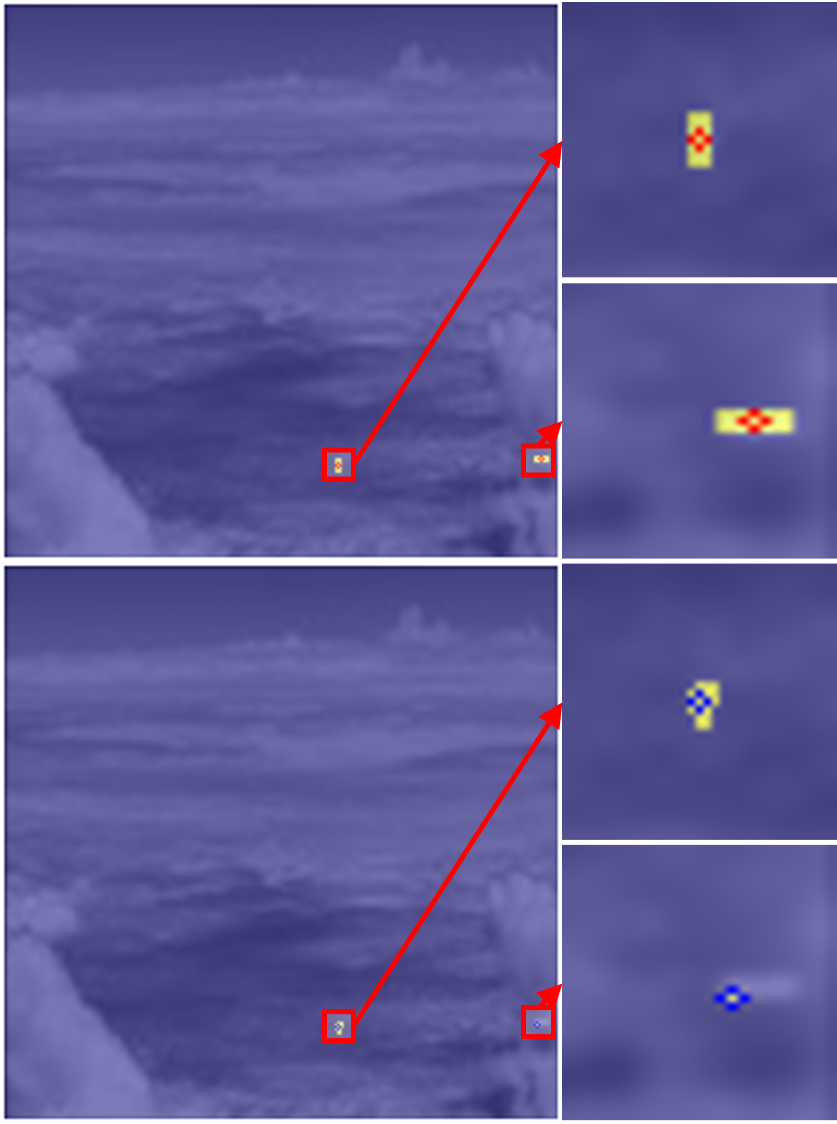}}
    \par
    \subfloat{\centering
        \resizebox{\linewidth}{!}{}}
    \caption{Comparison of different metrics.
        \textcolor{red}{Red} and \textcolor{blue}{blue} boxes to highlight the target regions.
        \textcolor{red}{Red} and \textcolor{blue}{blue} points indicate the target centroids in ground truth (GT) masks and predictions, respectively.
        Zoom in on the digital color version for details.
        ``Ori.''~\cite{IRSTD-DNANet} and ``OPDC'' refer to the original distance-based strategy and the proposed OPDC strategy for target matching.
    }
    \label{fig:total_metric_comparison}
\end{figure*}

\begin{figure*}[!h]
    \centering
    \includegraphics[width=\linewidth]{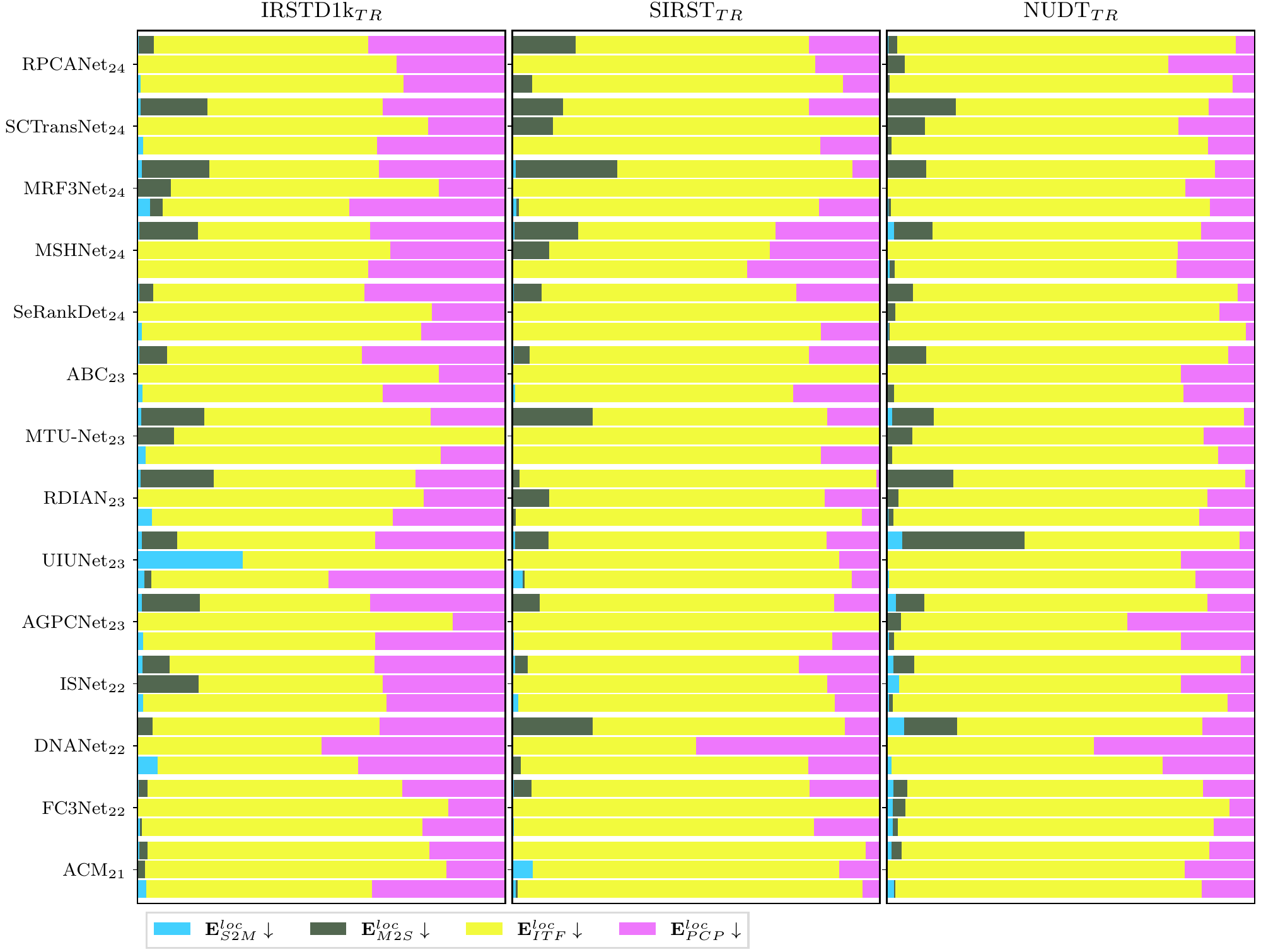}
    \caption{Cross-dataset localization error ratios corresponding to IRSTD1k$_{TE}$, SIRST$_{TE}$, and NUDT$_{TE}$ from bottom to top for the models trained on different datasets.}
    \label{fig:total_loc_error_ratio}
\end{figure*}

\begin{figure*}[!h]
    \centering
    \includegraphics[width=\linewidth]{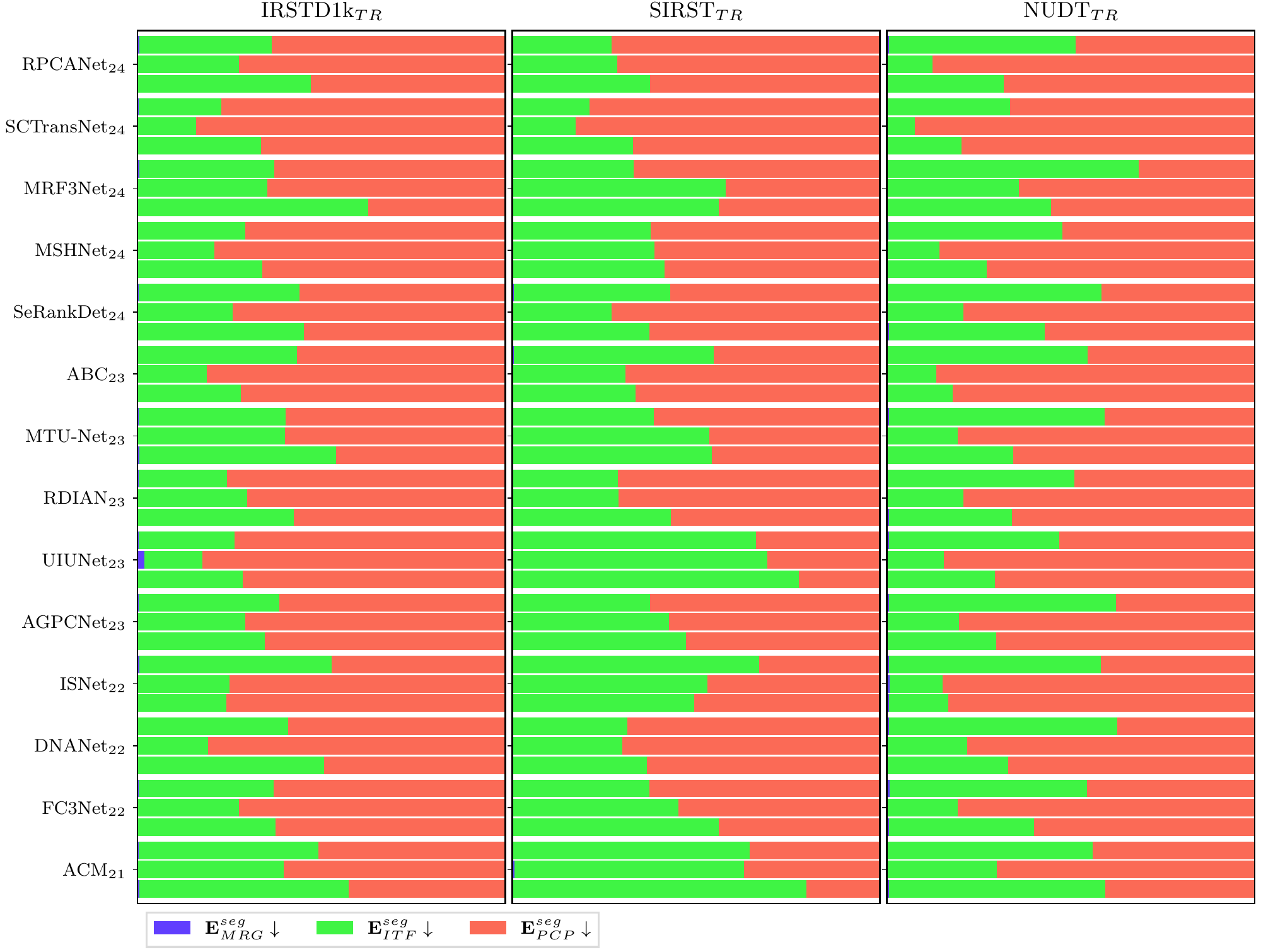}
    \caption{Cross-dataset segmentation error ratios corresponding to IRSTD1k$_{TE}$, SIRST$_{TE}$, and NUDT$_{TE}$ from bottom to top for the models trained on different datasets.}
    \label{fig:total_seg_error_ratio}
\end{figure*}

This section further adds the following:
\begin{itemize}[leftmargin=*,itemsep=0em,topsep=-1em,parsep=0em]
    \item Fig.~\ref{fig:total_metric_comparison}: Clearer figure visualization for improved readability.
    \item Tab.~\ref{tab:total_error_datasets}: Complete cross-dataset error analysis table (only the first section is presented in the main text due to space constraints).
    \item Fig.~\ref{fig:total_loc_error_ratio} and Fig.~\ref{fig:total_seg_error_ratio}: Clearer statistical visualization of cross-dataset error ratios.
\end{itemize}

\begin{table*}[!h]
    \centering
    \caption{Cross-dataset error analysis for the models trained on different datasets.}
    \label{tab:total_error_datasets}
    \resizebox{\linewidth}{!}{
\begin{tabular}{rr|rrrrrrrrrrrrrr}
    \toprule[2pt]
     &
     & ACM$_{21}$~\cite{IRSTD-ACM-nIoU}
     & FC3Net$_{22}$~\cite{IRSTD-FC3Net}
     & DNANet$_{22}$~\cite{IRSTD-DNANet}
     & ISNet$_{22}$~\cite{IRSTD-ISNet}
     & AGPCNet$_{23}$~\cite{IRSTD-AGPCNet}
     & UIUNet$_{23}$~\cite{IRSTD-UIUNet}
     & RDIAN$_{23}$~\cite{IRSTD-RDIAN}
     & MTU-Net$_{23}$~\cite{IRSTD-MTU-Net}
     & ABC$_{23}$~\cite{IRSTD-ABC}
     & SeRankDet$_{24}$~\cite{IRSTD-SeRankDet}
     & MSHNet$_{24}$~\cite{IRSTD-MSHNet}
     & MRF3Net$_{24}$~\cite{IRSTD-MRF3Net}
     & SCTransNet$_{24}$~\cite{IRSTD-SCTransNet}
     & RPCANet$_{24}$~\cite{IRSTD-RPCANet}                                                                                                                                                                               \\
    \midrule[1pt]
    \multicolumn{16}{r}{\textit{\textbf{Trained on IRSTD1k$_{TR}$~\cite{IRSTD-ISNet}.}}}                                                                                                                                \\
    \midrule[1pt]
    \multirow{9}{*}{\rotatebox{90}{IRSTD1k$_{TE}$~\cite{IRSTD-ISNet}}}
     & IoU$^{seg}_{pix}\uparrow$                 & 5.387e-01 & 6.172e-01 & 6.694e-01 & 5.476e-01 & 6.139e-01 & 6.317e-01 & 6.529e-01 & 6.543e-01 & 6.361e-01 & 6.755e-01 & 6.497e-01 & 6.742e-01 & 6.610e-01 & 6.607e-01 \\
     & $\mathbf{E}^{seg}_{MRG}\downarrow$        & 2.132e-03 & 4.340e-04 & 5.610e-04 & 4.830e-04 & 4.340e-04 & 4.440e-04 & 5.410e-04 & 4.580e-04 & 4.680e-04 & 4.250e-04 & 0.000e+00 & 4.420e-04 & 4.770e-04 & 4.570e-04 \\
     & $\mathbf{E}^{seg}_{ITF}\downarrow$        & 2.630e-01 & 1.431e-01 & 1.669e-01 & 1.089e-01 & 1.330e-01 & 1.048e-01 & 1.470e-01 & 1.855e-01 & 1.017e-01 & 1.463e-01 & 1.186e-01 & 2.039e-01 & 1.135e-01 & 1.595e-01 \\
\end{tabular}
}
\end{table*}

\section{Synthetic Data Experiment}
\label{supp:synthetic_data}

\begin{figure*}[!h]
    \centering
    \subfloat[Sample 1 (Original).]{\label{fig:original_sample_1}
        \centering
        \includegraphics[width=0.49\linewidth]{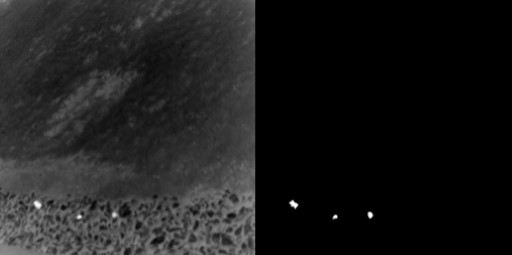}
    }
    \subfloat[Sample 1 (Augmented).]{\label{fig:augmented_sample_1}
        \centering
        \includegraphics[width=0.49\linewidth]{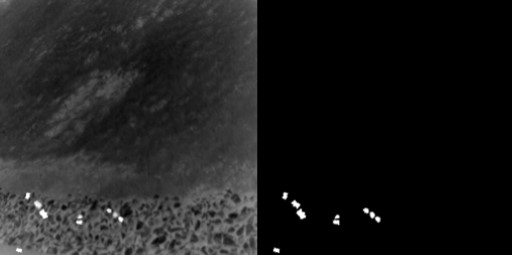}
    }
    \par
    \subfloat[Sample 2 (Original).]{\label{fig:original_sample_2}
        \centering
        \includegraphics[width=0.49\linewidth]{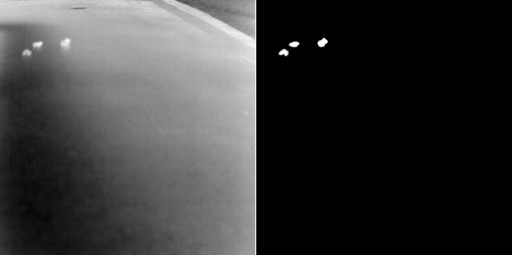}
    }
    \subfloat[Sample 2 (Augmented).]{\label{fig:augmented_sample_2}
        \centering
        \includegraphics[width=0.49\linewidth]{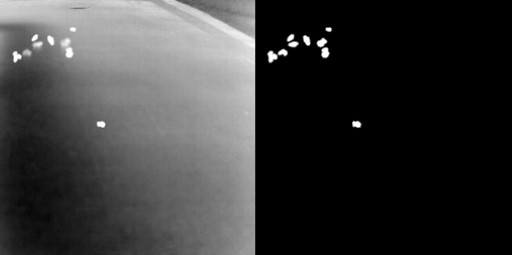}
    }
    \caption{Visual demonstration of sample augmentation effects.}
    \label{fig:samples_aug}
    \vspace{-1ex}
\end{figure*}

To further investigate algorithm performance in complex scenarios with with densely distributed targets, we augment the testset of the existing IRSTD1k~\cite{IRSTD-ISNet} using a randomized copy-paste augmentation strategy and construct the synthetic dataset, \ie, IRSTD1k$^{AUG}_{TE}$.
For copyright compliance, we exclusively performed data augmentation on the IRSTD1k dataset (\href{https://github.com/RuiZhang97/ISNet}{MIT License}), the only benchmark in our study with explicit redistribution permissions.
All synthetic data will be publicly released under the same license terms (MIT License), providing legally compliant yet challenging test cases that faithfully extend the benchmark's inherent properties.

The strategy intelligently generates new target instances by:
\begin{enumerate}[leftmargin=*,itemsep=0em,topsep=0em,parsep=0em]
    \item Extract valid target regions from the ground-truth mask using connected component analysis.
    \item Select targets for replication based on area-weighted sampling, prioritizing smaller targets.
    \item Generate copies at $2\times$ the original target count, with a maximum of 7 new targets per image.
    \item Augment each copy with random transformations including scaling ($0.5\text{-}1.1\times$), rotation ($0^\circ\text{-}180^\circ$), and positional perturbations ($75\%$ near original targets and $25\%$ globally distributed).
\end{enumerate}
This approach realistically increases object density and morphological diversity, expanding the original dataset while maintaining scene coherence, thereby creating more challenging test conditions for evaluating algorithm robustness.
And some samples are shown in Fig.~\ref{fig:samples_aug}.

\begin{figure*}[!h]
    \centering
    \subfloat[Localization errors.]{\label{fig:total_loc_error_ratio_aug}
        \centering
        \includegraphics[width=\linewidth]{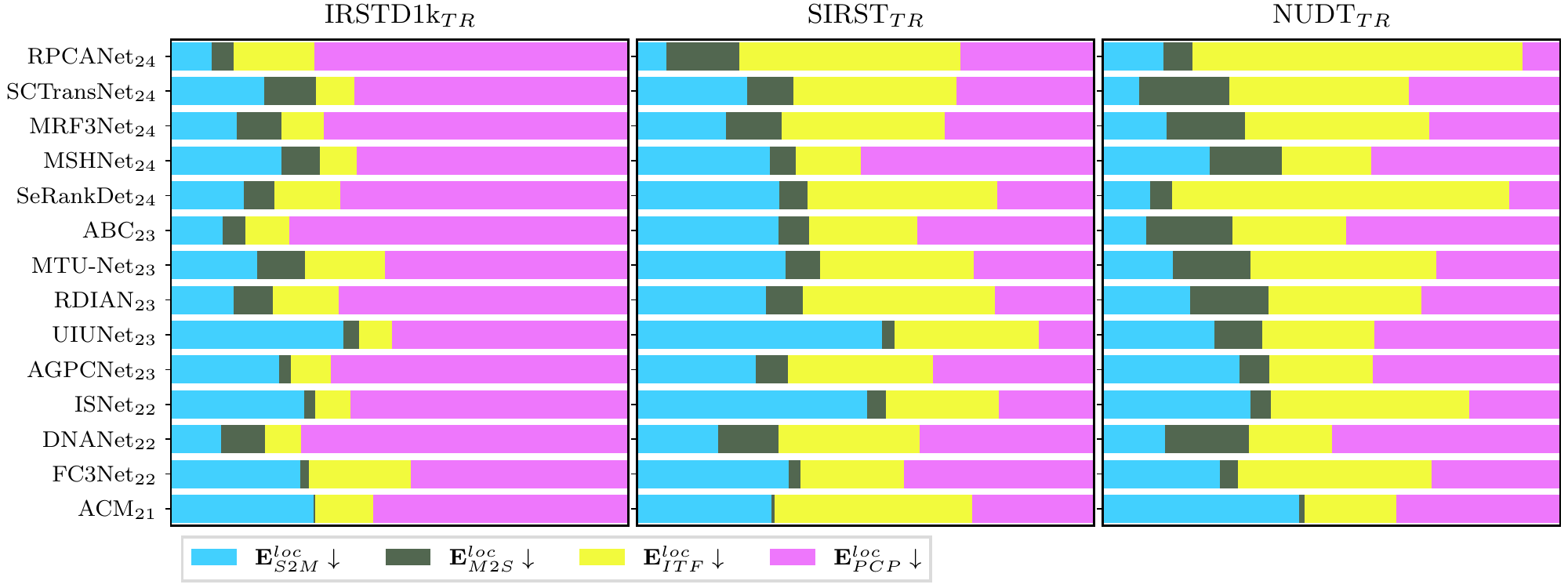}
    }
    \par
    \subfloat[Segmentation errors.]{\label{fig:total_seg_error_ratio_aug}
        \centering
        \includegraphics[width=\linewidth]{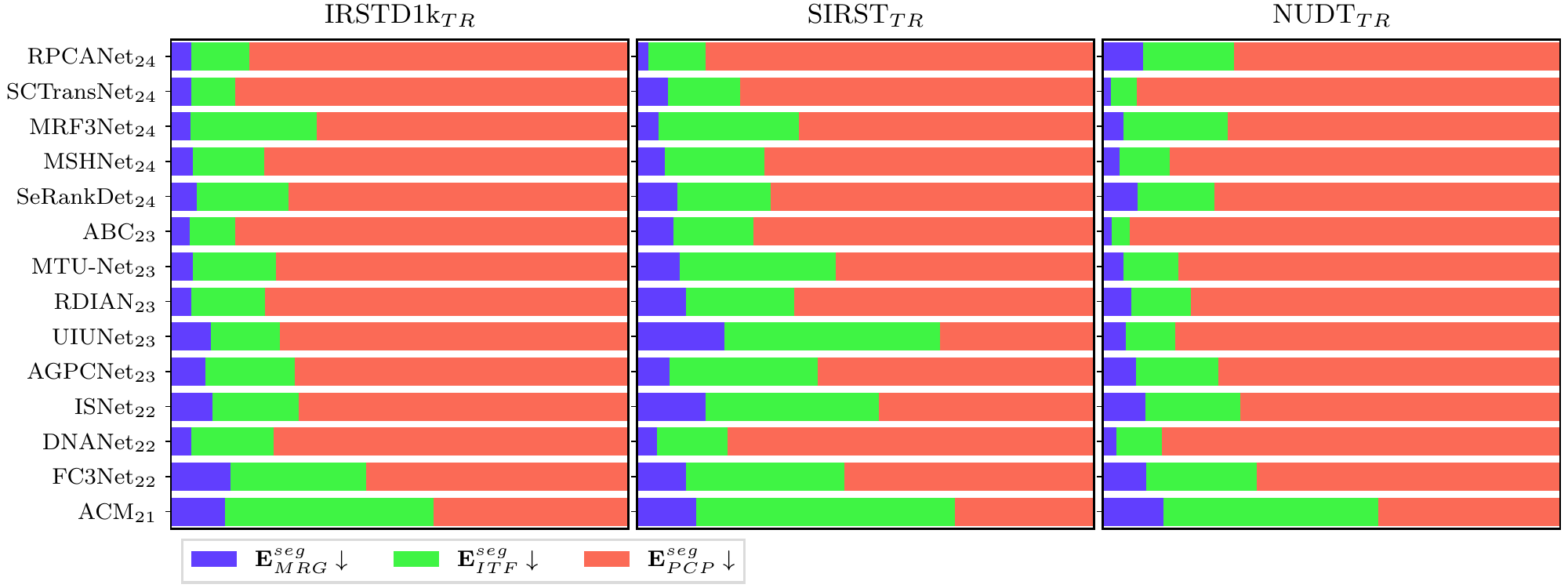}
    }
    \caption{Cross-dataset error ratios on the synthetic dataset IRSTD1k$^{AUG}_{TE}$ for the models trained on different datasets.}
    \label{fig:error_ratio_aug}
    \vspace{-1ex}
\end{figure*}

As evidenced in Tab~\ref{tab:metric_error_all_datasets_aug}, existing models exhibit significant performance degradation on this new challenging dataset, with our error statistics showing marked increases across all error types.
Fig.~\ref{fig:error_ratio_aug} provides a visual breakdown of how different error subtypes contribute to the overall performance decline.
Notably, when comparing Fig.~\ref{fig:total_loc_error_ratio_aug} with Fig.~\ref{fig:total_loc_error_ratio}, we observe a substantial increase in the proportion of $\mathbf{E}^{loc}_{S2M}$, which directly validates our analysis about it in Sec.~\ref{sec:error_analysis_method} of the main text.
Similarly, the dramatic rise in $\mathbf{E}^{seg}_{MRG}$ shown in Fig.~\ref{fig:total_seg_error_ratio_aug} versus Fig.~\ref{fig:total_seg_error_ratio} indicates these models struggle to effectively distinguish between individual target instances in high-density target distributions.

\begin{table*}[!h]
    \centering
    \caption{Performance and error analysis on the synthetic dataset IRSTD1k$^{AUG}_{TE}$ which utilize data augmentation strategies to significantly expand the number and morphology of targets on each image.
        Colors \first{red}, \second{green} and \third{blue} represent the first, second and third ranked results.}
    \label{tab:metric_error_all_datasets_aug}
    \resizebox{\linewidth}{!}{
\begin{tabular}{l|rrrrrrrrrrrrrr}
    \toprule[2pt]
                                       & ACM$_{21}$~\cite{IRSTD-ACM-nIoU}
                                       & FC3Net$_{22}$~\cite{IRSTD-FC3Net}
                                       & DNANet$_{22}$~\cite{IRSTD-DNANet}
                                       & ISNet$_{22}$~\cite{IRSTD-ISNet}
                                       & AGPCNet$_{23}$~\cite{IRSTD-AGPCNet}
                                       & UIUNet$_{23}$~\cite{IRSTD-UIUNet}
                                       & RDIAN$_{23}$~\cite{IRSTD-RDIAN}
                                       & MTU-Net$_{23}$~\cite{IRSTD-MTU-Net}
                                       & ABC$_{23}$~\cite{IRSTD-ABC}
                                       & SeRankDet$_{24}$~\cite{IRSTD-SeRankDet}
                                       & MSHNet$_{24}$~\cite{IRSTD-MSHNet}
                                       & MRF3Net$_{24}$~\cite{IRSTD-MRF3Net}
                                       & SCTransNet$_{24}$~\cite{IRSTD-SCTransNet}
                                       & RPCANet$_{24}$~\cite{IRSTD-RPCANet}                                                                                                                                                                                                                                      \\
\end{tabular}
}
\end{table*}

\section{Other Discussions}
\label{supp:other_discussions}

\subsection{Further Error Analysis}
\label{supp:sec:more_error_analysis}

\parhead{Significant Value of Error Analysis.}
The goal of our error analysis is not merely to present total performance but to provide a more fine-grained understanding of model behavior, exposing the limitations and failure modes that are often hidden behind aggregated performance metrics.
For instance, as demonstrated in Sec.~\ref{supp:synthetic_data}, Fig.~\ref{fig:error_ratio} and Fig.~\ref{fig:error_ratio_aug} present the relative proportions of error components, which help identify which types of error dominate under different conditions.
In dense-target scenarios, our analysis reveals that $\mathbf{E}^{seg}_{MRG}$ and $\mathbf{E}^{tgt}_{S2M}$ are the most significant contributors to failure, as discussed in Sec.~\ref{sec:error_analysis_method}.
Furthermore, Tab.~\ref{tab:total_error_datasets} provide quantitative error values and evaluate the severity of each error type in detail, enabling more targeted performance improvement strategies beyond overall score optimization.

\parhead{Potential Impacts of Model Structure.}
Different structural choices in the model design clearly correlate with distinct error distributions.
For instance, some models such as ACM~\cite{IRSTD-ACM-nIoU} and FC3Net~\cite{IRSTD-FC3Net}, which lack explicit global modeling or deep semantic suppression, exhibit high $\mathbf{E}^{seg}_{ITF}$ and $\mathbf{E}^{loc}_{ITF}$ errors, often confusing textured backgrounds as targets.
ISNet~\cite{IRSTD-ISNet} and UIUNet~\cite{IRSTD-UIUNet}, while integrating attention and multi-scale decoding, still suffer from structural inconsistencies and incorrect merging or splitting (\eg, high $\mathbf{E}^{seg}_{PCP}$ or $\mathbf{E}^{loc}_{S2M}$/$\mathbf{E}^{loc}_{M2S}$ errors), reflecting limitations of shallow local attention.
These observations highlight structural limitations and support the need for better context integration, stronger structural priors, and error-aware learning objectives.

\subsection{Why Use the Multiplicative Form?}
\label{supp:why_use_multiplicative_form}

We chose the multiplicative form $\text{hIoU} = \text{IoU}^{loc}_{tgt} \times \text{IoU}^{seg}_{pix}$ in Equ.~\ref{equ:hiou} over the additive alternative $\text{aIoU} = 0.5(\text{IoU}^{loc}_{tgt} + \text{IoU}^{seg}_{pix})$ for modeling the interdependence of localization and segmentation in IRSTD:
\begin{enumerate}[leftmargin=*,itemsep=0em,topsep=0em,parsep=0em]
    \item Since $0\le \text{IoU}_{loc}, \text{IoU}_{seg}\le1$, both $\text{hIoU}$ and $\text{aIoU}$ lie in $[0,1]$:
          \begin{enumerate}
              \item Both equal 1 only when both $\text{IoU}_{loc}=\text{IoU}_{seg}=1$.
                    $\text{hIoU}=0$ if either component is 0, but $\text{aIoU}=0$ only if both are 0.
              \item $\text{hIoU}=\text{IoU}_{loc} \text{IoU}_{seg}>t \Rightarrow \text{IoU}_{loc}=\frac{t}{\text{IoU}_{seg}}>t$ and $\text{IoU}_{seg}=\frac{t}{\text{IoU}_{loc}}>t$, so any threshold $t$ is enforced simultaneously.
                    By contrast, $\text{aIoU}=0.5(\text{IoU}_{loc}+\text{IoU}_{seg})>t \Rightarrow \text{IoU}_{loc}>2t-\text{IoU}_{seg}$ corresponds to the half-space, a much larger region that includes points where one coordinate can be far below $t$ (\eg, $(\text{IoU}_{loc}, \text{IoU}_{seg})=(2t-1, 1)$). Consequently, $\text{aIoU}$ covers a broader area, over-approximating joint performance, whereas hIoU strictly requires both components to exceed $t$.
          \end{enumerate}
    \item Both increase monotonically in each argument as follows:
          \begin{enumerate}
              \item $\frac{\partial \text{hIoU}}{\partial \text{IoU}_{loc}} = \text{IoU}_{seg}, \frac{\partial \text{hIoU}}{\partial \text{IoU}_{seg}} = \text{IoU}_{loc}$: The coupling means that when one component is low, the influence of the other is diminished, ensuring that isolated improvements cannot disproportionately boost the overall score.
              \item $\frac{\partial \text{aIoU}}{\partial \text{IoU}_{loc}} = \frac{\partial \text{aIoU}}{\partial \text{IoU}_{seg}} = 0.5$: In contrast, $\text{aIoU}$ assigns fixed, equal weight to each component, ignoring their interaction and thus diluting the impact of imbalance.
          \end{enumerate}
    \item The total error can be written as follows:
          \begin{enumerate}
              \item For $\text{hIoU}$: $\mathbf{E}_\text{hIoU} = 1-\text{hIoU} = 1-(1-\mathbf{E}^{loc})(1-\mathbf{E}^{seg}) = \mathbf{E}^{loc}+\mathbf{E}^{seg}-\mathbf{E}^{loc}\mathbf{E}^{seg} = \begin{cases}
                            \mathbf{E}^{seg}(1-\mathbf{E}^{loc})+\mathbf{E}^{loc}\ge \mathbf{E}^{loc} \\
                            \mathbf{E}^{loc}(1-\mathbf{E}^{seg})+\mathbf{E}^{seg}\ge \mathbf{E}^{seg}
                        \end{cases}
                        \Rightarrow \max(\mathbf{E}^{loc},\mathbf{E}^{seg})\le \mathbf{E}_\text{hIoU} \le \min(1, \mathbf{E}^{loc}+\mathbf{E}^{seg})$
              \item For aIoU: $\mathbf{E}_\text{aIoU} = 1-\text{aIoU} = 0.5(\mathbf{E}^{loc}+\mathbf{E}^{seg})$.
              \item $\mathbf{E}_\text{hIoU}$ intuitively reflects that ``the shortcoming dictates the performance ceiling''.
                    In contrast, $\mathbf{E}_\text{aIoU}$ is always $0.5(\mathbf{E}^{loc}+\mathbf{E}^{seg})$, lacking the shortcoming effect.
                    Thus, $\mathbf{E}_\text{hIoU}$ more faithfully captures the coupled relationship in IRSTD.
          \end{enumerate}
\end{enumerate}

The multiplicative form naturally penalizes any weak link, making it a more faithful, robust metric for IRSTD than the additive aIoU.
No high score can ``hide'' a bad component, whereas the additive form can still report misleadingly high scores even if one factor is near zero, thus failing to reflect genuine end-to-end performance.

\subsection{Why Use the IoU-based Form?}
\label{supp:why_use_iou_based_form}

Our choice to adopt an IoU-based formulation is intentional and motivated by both practical and conceptual considerations.
\begin{enumerate}[leftmargin=*,itemsep=0em,topsep=0em,parsep=0em]
    \item Using the IoU form ensures consistency with the segmentation IoU, thereby yielding a uniform metric structure across both pixel and target levels.
          This consistency simplifies the interpretation and facilitates a coherent hierarchical error decomposition framework.
          Specifically:
          \begin{enumerate}
              \item It enables pixel-level and target-level components to have the same variation trends and intrinsic meanings.
              \item It allows a unified performance and error analysis principle based on set intersection-over-union, to be applied at different levels.
              \item It can further simplify the final computation of the hIoU, as the term $\sum_{i=1}^{K} |\text{TP}^{[i]}_{tgt}|$ can be cleanly canceled out due to structural consistency.
          \end{enumerate}
    \item IoU has the added advantage of directly reflecting spatial overlap, which is particularly meaningful for tasks involving localization and segmentation.
          This makes it more intuitive and analytically convenient in the context of IRSTD.
\end{enumerate}

\subsection{Extended Discussion on F1-score}
\label{supp:target_level_f1_score}

\parhead{Target-level F1-score (F1$_{tgt}$).}
As stated in the main text, the F1 we used follows the pixel-level formulation that has been widely adopted in IRSTD~\cite{IRSTD-SCTransNet}.
Therefore, our implementation of F1 is aligned with the precedent set by prior literature.
F1/precision/recall can also be formulated at the target level.
Notably, the recall in this context corresponds to the commonly used target-level IRSTD metric Pd.
To provide a more comprehensive evaluation, we additionally supplement the target-level F1$_{tgt}$ in Tab.~\ref{tab:target_level_f1_score}.

\begin{table*}[!t]
    \centering
    \caption{Cross-dataset performance analysis. Colors \first{red}, \second{green} and \third{blue} represent the first, second and third ranked results. \textbf{Besides the reported metrics, the target-level F1-score (F1$_{tgt}$) is included in the content of Tab.~\ref{tab:metric_all_datasets}.}}
    \label{tab:target_level_f1_score}
    \resizebox{\linewidth}{!}{
\begin{tabular}{ll|rrrrrrrrrrrrrr}
    \toprule[2pt]
                                 &
                                 & ACM$_{21}$~\cite{IRSTD-ACM-nIoU}
                                 & FC3Net$_{22}$~\cite{IRSTD-FC3Net}
                                 & DNANet$_{22}$~\cite{IRSTD-DNANet}
                                 & ISNet$_{22}$~\cite{IRSTD-ISNet}
                                 & AGPCNet$_{23}$~\cite{IRSTD-AGPCNet}
                                 & UIUNet$_{23}$~\cite{IRSTD-UIUNet}
                                 & RDIAN$_{23}$~\cite{IRSTD-RDIAN}
                                 & MTU-Net$_{23}$~\cite{IRSTD-MTU-Net}
                                 & ABC$_{23}$~\cite{IRSTD-ABC}
                                 & SeRankDet$_{24}$~\cite{IRSTD-SeRankDet}
                                 & MSHNet$_{24}$~\cite{IRSTD-MSHNet}
                                 & MRF3Net$_{24}$~\cite{IRSTD-MRF3Net}
                                 & SCTransNet$_{24}$~\cite{IRSTD-SCTransNet}
                                 & RPCANet$_{24}$~\cite{IRSTD-RPCANet}                                                                                                                                                    \\
    \midrule[0.5pt]
    \multicolumn{2}{c|}{Params.} & 0.5M
                                 & 6.9M
                                 & 4.7M
                                 & 1.1M
                                 & 12.4M
                                 & 50.5M
                                 & 0.1M
                                 & 4.1M
                                 & 73.5M
                                 & 108.9M
                                 & 4.1M
                                 & 0.5M
                                 & 11.2M
                                 & 0.7M                                                                                                                                                                                   \\
\end{tabular}
}
\end{table*}

\parhead{Why not $\text{F1}_{tgt} \times \text{nIoU}_{pix}$?}
There are fundamental differences in the way they handle error attribution.
The product of target-level $\text{F1}_{tgt}$ and pixel-level $\text{nIoU}_{pix}$ can lead to redundant penalization, as it implicitly assumes independence between segmentation and localization errors.
However, the two types of errors are often correlated, for instance, inaccurate localization will simultaneously degrade whole segmentation performance.
As a result, this form tends to double-count the impact of shared failure sources, and thus does not faithfully reflect the overall performance.
In contrast, our hIoU is designed to disentangle these two layers of performance.
Localization quality is measured first, and any performance loss due to missed or poorly localized targets is entirely attributed to the target-level.
Segmentation quality is then evaluated only within the region of correctly matched targets, focusing solely on the spatial overlap and avoiding entanglement with localization failure.
This design ensures that each layer is evaluated in a complementary manner and errors are attributed unambiguously to their true source.
Consequently, hIoU offers a more accurate, interpretable, and fair assessment of the model's overall performance, avoiding the distortion introduced by overlapping error contributions.

\subsection{Multi-Frame Infrared Small Target Detection}
\label{supp:sec:multi_frame_infrared_small_target_detection}

Some advances leverage temporal cues (\ie, multi-frame IRSTD) to enhance robustness against clutter, including motion direction encoding~\cite{MIRSTD-DTUM} and recurrent refinement with motion compensation~\cite{MIRSTD-RFR}, effectively exploiting spatiotemporal dynamics.
Recent work~\cite{MIRSTD-DeepPro} explores the temporal-profile perspective by reformulating detection as a one-dimensional anomaly task, offering high efficiency.
These works highlight the emerging shift from purely spatial designs toward spatio-temporal paradigms.

Our evaluation framework is agnostic to the detection paradigm and is designed to be equally applicable to both single-frame and multi-frame IRSTD methods.
In particular, the proposed metrics and analysis mechanisms remain consistent across these settings without any need for structural adaptation.
In Tab.~\ref{tab:multiframe_results}, we conduct experiments on a representative multi-frame dataset~\cite{MIRSTD-DTUM} and evaluate several state-of-the-art multi-frame IRSTD methods~\cite{MIRSTD-DTUM,MIRSTD-RFR,MIRSTD-DeepPro} using our proposed framework.

\begin{table*}[!h]
    \centering
    \caption{Results of multi-frame IRSTD methods.}
    \label{tab:multiframe_results}
    \resizebox{0.6\linewidth}{!}{
        \begin{tabular}{l|cccccccccc}
            \toprule[2pt]
                                               &
            \myhiou                            &
            IoU$^{seg}_{pix}\uparrow$          &
            $\mathbf{E}^{seg}_{MRG}\downarrow$ &
            $\mathbf{E}^{seg}_{ITF}\downarrow$ &
            $\mathbf{E}^{seg}_{PCP}\downarrow$ &
            IoU$^{loc}_{tgt}\uparrow$          &
            $\mathbf{E}^{loc}_{S2M}\downarrow$ &
            $\mathbf{E}^{loc}_{M2S}\downarrow$ &
            $\mathbf{E}^{loc}_{ITF}\downarrow$ &
            $\mathbf{E}^{loc}_{PCP}\downarrow$                                                                                 \\
            \midrule[1pt]
            \cite{MIRSTD-DTUM}                 & 0.552 & 0.832 & 0.000 & 0.107 & 0.061 & 0.683 & 0.000 & 0.010 & 0.189 & 0.119 \\
            \cite{MIRSTD-RFR}                  & 0.546 & 0.816 & 0.000 & 0.117 & 0.067 & 0.669 & 0.000 & 0.012 & 0.195 & 0.123 \\
            \cite{MIRSTD-DeepPro}              & 0.591 & 0.822 & 0.000 & 0.114 & 0.064 & 0.719 & 0.000 & 0.006 & 0.151 & 0.124 \\
            \bottomrule[2pt]
        \end{tabular}
    }
\end{table*}

\subsection{Validation under Occlusion, Deformation, and Connectivity}
\label{sec:validation_occlusion_deformation_connectivity}

Our work focuses on the IRSTD task, where the targets often exhibit significant boundary ambiguity and shape diversity, as shown in Fig.~\ref{fig:total_metric_comparison} and Fig.~\ref{fig:samples_aug}.
Our OPDC strategy does not rely on the content of the input image (\eg, weather conditions or image quality), but instead operates solely on the binary prediction and ground-truth masks.
It incorporates both distance-based and region-overlap constraints, which helps reduce interference caused by target occlusion, deformation, or connectivity.
We also supplement experiments of OPDC under these challenging conditions by creating three data subsets with random target occlusion, deformation, and connectivity.
The proposed OPDC strategy is compared with the original distance-based approach~\cite{IRSTD-DNANet}.
The ratio of predefined target pairs that are successfully matched is summarized in Tab.~\ref{tab:predicted_matching_success_rate}.
This experiment demonstrates that our OPDC performs better than the distance-based method~\cite{IRSTD-DNANet}.

\begin{table*}[!h]
    \centering
    \caption{Success rate of predefined target pair matching.}
    \label{tab:predicted_matching_success_rate}
    \resizebox{0.3\linewidth}{!}{
        \begin{tabular}{l|ccc}
            \toprule[2pt]
                     & Occlusion & Deformation & Connectivity \\
            \midrule[1pt]
            OPDC     & 1.000     & 1.000       & 0.949        \\
            Distance & 0.420     & 0.320       & 0.379        \\
            \bottomrule[2pt]
        \end{tabular}
    }
\end{table*}

\subsection{Validation on Medical Small Object Detection}
\label{supp:sec:validation_on_medical_small_object_detection}

To verify its applicability, we transfer the proposed framework to the medical image domain, specifically, the polyp segmentation task, which also involves small and irregular targets.
We evaluate the classic method~\cite{PraNet} in Tab.~\ref{tab:medical_small_object_detection} using our framework and observe clear localization-related errors, including large values in $\mathbf{E}^{loc}_{ITF}$ and $\mathbf{E}^{loc}_{PCP}$, corresponding to missed detections and incorrect background predictions.
These issues are also visually evident in its predictions, confirming that established methods can also suffer from target-level limitations not reflected in traditional metrics.
This extension further shows the generality and diagnostic value of our framework beyond IRSTD.

\begin{table*}[!h]
    \centering
    \caption{Experiments on medical small object detection, \ie, the polyp segmentation task.}
    \label{tab:medical_small_object_detection}
    \resizebox{0.6\linewidth}{!}{
        \begin{tabular}{l|cccccccccc}
            \toprule[2pt]
                                               &
            \myhiou                            &
            IoU$^{seg}_{pix}\uparrow$          &
            $\mathbf{E}^{seg}_{MRG}\downarrow$ &
            $\mathbf{E}^{seg}_{ITF}\downarrow$ &
            $\mathbf{E}^{seg}_{PCP}\downarrow$ &
            IoU$^{loc}_{tgt}\uparrow$          &
            $\mathbf{E}^{loc}_{S2M}\downarrow$ &
            $\mathbf{E}^{loc}_{M2S}\downarrow$ &
            $\mathbf{E}^{loc}_{ITF}\downarrow$ &
            $\mathbf{E}^{loc}_{PCP}\downarrow$   \\
            \midrule[1pt]
            \cite{PraNet}                      &
            0.589                              &
            0.881                              &
            0.003                              &
            0.050                              &
            0.066                              &
            0.669                              &
            0.000                              &
            0.000                              &
            0.143                              &
            0.188                                \\
            \bottomrule[2pt]
        \end{tabular}
    }
\end{table*}

\subsection{Validation on Different Target Attributes}
\label{sec:validation_on_different_target_attributes}

Since our metrics are computed based solely on binary prediction and ground truth masks, they do not rely on the input image itself and the target contrast in the image has no influence on the evaluation process.
To analyze the influence of target size and density attributes, we manually adjust the size and spatial density of targets and the results are reported in Tab.~\ref{tab:different_target_attributes}.
The results show that while the metrics are generally stable, changes in target size or density do lead to observable variations in hIoU.
This behavior is expected and reasonable: modifying target size or density can alter the relative distances and overlaps between targets, which directly affects localization accuracy, segmentation overlap, and ultimately the joint hIoU score.

\begin{table*}[!h]
    \centering
    \caption{Experiments on different target attributes, including size and density.}
    \label{tab:different_target_attributes}
    \resizebox{0.7\linewidth}{!}{
        \begin{tabular}{l|cccccccccc}
            \toprule[2pt]
                                               &
            \myhiou                            &
            IoU$^{seg}_{pix}\uparrow$          &
            $\mathbf{E}^{seg}_{MRG}\downarrow$ &
            $\mathbf{E}^{seg}_{ITF}\downarrow$ &
            $\mathbf{E}^{seg}_{PCP}\downarrow$ &
            IoU$^{loc}_{tgt}\uparrow$          &
            $\mathbf{E}^{loc}_{S2M}\downarrow$ &
            $\mathbf{E}^{loc}_{M2S}\downarrow$ &
            $\mathbf{E}^{loc}_{ITF}\downarrow$ &
            $\mathbf{E}^{loc}_{PCP}\downarrow$                                                                                 \\
            \midrule[1pt]
            Original                           & 0.430 & 0.632 & 0.000 & 0.182 & 0.186 & 0.687 & 0.000 & 0.000 & 0.207 & 0.106 \\
            Smaller (Erosion)                  & 0.422 & 0.614 & 0.002 & 0.192 & 0.192 & 0.686 & 0.000 & 0.000 & 0.206 & 0.108 \\
            Larger (Dilation)                  & 0.437 & 0.633 & 0.001 & 0.180 & 0.186 & 0.690 & 0.000 & 0.000 & 0.209 & 0.101 \\
            Sparser                            & 0.424 & 0.622 & 0.001 & 0.188 & 0.189 & 0.682 & 0.000 & 0.000 & 0.208 & 0.110 \\
            Denser                             & 0.441 & 0.634 & 0.000 & 0.176 & 0.190 & 0.696 & 0.000 & 0.000 & 0.207 & 0.097 \\
            \bottomrule[2pt]
        \end{tabular}
    }
\end{table*}

In addition, we conduct experiments in Sec.~\ref{supp:synthetic_data} by creating synthetic scenarios with more and denser small targets.
As shown in the comparison between the original dataset (Tab.~\ref{tab:metric_all_datasets}) and the synthetic dataset (Tab.~\ref{tab:metric_error_all_datasets_aug}), the average differences between other metrics and hIoU are listed as Tab.~\ref{tab:metric_differences}.
These results show that, except for Pd and Fa, the performance gap between hIoU and other metrics remains relatively stable.
The changes in Pd and Fa may be related to their strong dependence on target matching, which is itself sensitive to variations in target number and density.

\begin{table*}[!h]
    \centering
    \caption{Average differences between other metrics and hIoU.}
    \label{tab:metric_differences}
    \resizebox{0.4\linewidth}{!}{
        \begin{tabular}{ccccccc}
            \toprule[2pt]
            \myiou          &
            \myniou         &
            \myfone         &
            \mypd           &
            +OPDC$\uparrow$ &
            \myfa           &
            +OPDC$\uparrow$                                                      \\
            \midrule[1pt]
            -0.003          & 0.014 & 0.006 & -0.195 & -0.142 & 58.519 & -28.424 \\
            \bottomrule[2pt]
        \end{tabular}
    }
\end{table*}

\subsection{Experiments on NUDT-SIRST-Sea}
\label{supp:nudt_sirst_sea}

The selected datasets in the main text are the most commonly-used benchmarks in IRSTD.
As shown in Tab.~\ref{tab:metric_all_datasets}, existing methods still exhibit large performance variations across these datasets, suggesting inherent distribution differences in data.
We additionally introduce the NUDT-SIRST-Sea dataset~\cite{IRSTD-MTU-Net}, which features space-based infrared imagery of tiny ships and drastically different with existing data.
Results of RPCANet~\cite{IRSTD-RPCANet} on four datasets as listed in Tab.~\ref{tab:nudt_sirst_sea_results} show a significant performance drop under this distribution shift, confirming both the sensitivity of current methods and the usefulness of our framework in revealing such robustness issues.

\begin{table*}[!h]
    \centering
    \caption{Results of RPCANet~\cite{IRSTD-RPCANet} on four datasets.}
    \label{tab:nudt_sirst_sea_results}
    \resizebox{0.45\linewidth}{!}{
        \begin{tabular}{ccccccc}
            \toprule[2pt]
                    & IRSTD1k$_{TE}$~\cite{IRSTD-ISNet} & NUDT$_{TE}$~\cite{IRSTD-DNANet} & SIRST$_{TE}$~\cite{IRSTD-ACM-nIoU} & NUDT-SIRST-Sea$_{TE}$~\cite{IRSTD-MTU-Net} \\
            \midrule[1pt]
            \myiou  & 0.608                             & 0.291                           & 0.543                              & 0.001                                      \\
            \myfone & 0.756                             & 0.451                           & 0.704                              & 0.001                                      \\
            \mypd   & 0.886                             & 0.701                           & 0.927                              & 0.303                                      \\
            \myfa   & 28.145                            & 190.989                         & 124.066                            & 56666.629                                  \\
            \myhiou & 0.470                             & 0.344                           & 0.598                              & 0.017                                      \\
            \bottomrule[2pt]
        \end{tabular}
    }
\end{table*}

\begin{table*}[!h]
    \centering
    \caption{Attribute statistics for four datasets.}
    \label{tab:attribute_statistics}
    \resizebox{0.95\linewidth}{!}{
        \begin{tabular}{c|cccc|cccc}
            \toprule[2pt]
                                                       & \multicolumn{4}{c|}{Image Attributes} & \multicolumn{4}{c}{Target Attributes}                                                                                                                  \\
                                                       & Brightness                            & Brightness                            & Contrast Root Mean & Laplacian-based  & Average      & Average     & Target-Background & Foreground-Background \\
                                                       & Mean                                  & Standard Deviation                    & Square Contrast    & Noise Estimation & Target Count & Target Size & Contrast          & Area Ratio            \\
            \midrule[1pt]
            IRSTD1k$_{TE}$~\cite{IRSTD-ISNet}          & 0.344                                 & 0.149                                 & 0.149              & 43.436           & 1.477        & 51.148      & 131.865           & 1.786$\times 10^4$    \\
            SIRST$_{TE}$~\cite{IRSTD-ACM-nIoU}         & 0.428                                 & 0.098                                 & 0.098              & 40.068           & 1.267        & 30.289      & 1.936             & 3.795$\times 10^4$    \\
            NUDT$_{TE}$~\cite{IRSTD-DNANet}            & 0.419                                 & 0.127                                 & 0.127              & 110.505          & 1.426        & 34.703      & 1.525             & 4.263$\times 10^4$    \\
            NUDT-SIRST-Sea$_{TE}$~\cite{IRSTD-MTU-Net} & 0.244                                 & 0.076                                 & 0.076              & 66.729           & 2.238        & 10.011      & 0.881             & 1.761$\times 10^4$    \\
            \bottomrule[2pt]
        \end{tabular}
    }
\end{table*}

\begin{figure*}[!h]
    \centering
    \includegraphics[width=0.5\linewidth]{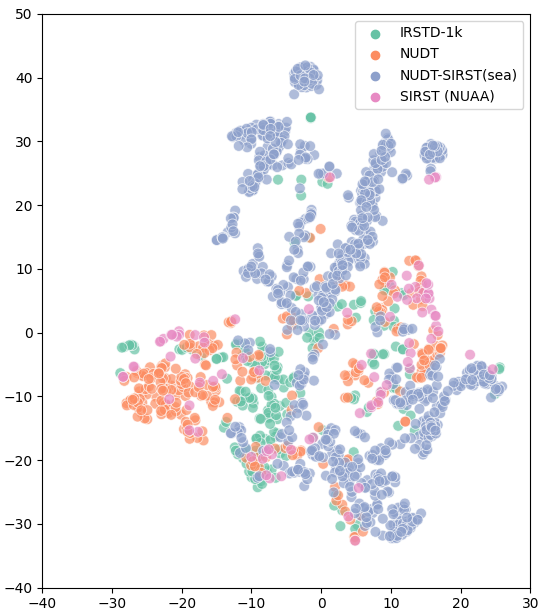}
    \caption{Illustration of attribute distribution for four datasets.}
    \label{fig:tsne_analysis}
\end{figure*}

To better characterize the differences in data distributions, we analyze and compare the four testing datasets, \ie, IRSTD1k$_{TE}$~\cite{IRSTD-ISNet}, SIRST$_{TE}$~\cite{IRSTD-ACM-nIoU}, NUDT$_{TE}$~\cite{IRSTD-DNANet}, and NUDT-SIRST-Sea$_{TE}$~\cite{IRSTD-MTU-Net}.
A set of hand-crafted statistical features from each image, including image attributes (\ie, brightness, contrast, and noise estimation) and target attributes (\ie, count, size, contrast, and area ratio) are extracted and summarized in Tab.~\ref{tab:attribute_statistics}.
They can be used to visualize the overall data distribution of these datasets based on t-SNE, as shown in Fig.~\ref{fig:tsne_analysis}.
The visualization offers an interpretable and model-agnostic way to clear distribution differences through images and target attributes.
While partial overlaps exist, each dataset occupies distinct regions, highlighting cross-dataset heterogeneity.
This underscores the necessity of cross-dataset analysis to avoid bias and improve generalization in the IRSTD research.

\section{Limitations and Future Extensions}
\label{supp:limitations_and_future_extensions}

\parhead{Broader Data and Application Scenarios.}
Given its status as a representative, mature, and challenging task with extensive research and diverse real-world applications, we selected IRSTD (Infrared Small Target Detection) as our primary focus.
Our work addresses critical limitations in its existing evaluation framework through targeted enhancements.
While this work provides focused analysis of 14 representative deep learning-based IRSTD methods~\cite{IRSTD-ACM-nIoU,IRSTD-FC3Net,IRSTD-DNANet,IRSTD-ISNet,IRSTD-AGPCNet,IRSTD-UIUNet,IRSTD-RDIAN,IRSTD-MTU-Net,IRSTD-ABC,IRSTD-SeRankDet,IRSTD-MSHNet,IRSTD-MRF3Net,IRSTD-SCTransNet,IRSTD-RPCANet}, its scope is fundamentally limited by relying exclusively on commonly-used infrared-only datasets (SIRST~\cite{IRSTD-ACM-nIoU}, IRSTD1k~\cite{IRSTD-ISNet}, and NUDT~\cite{IRSTD-DNANet}).
We also explore more challenging scenarios using synthetic data constructed based on data augmentation techniques in Sec.~\ref{supp:synthetic_data}.
However, given the generalizability of our focal problem and the model- and data-agnostic nature of the proposed analysis framework, it can be readily extended to broader data and application scenarios, such as the multi-frame IRSTD setting (Sec.~\ref{supp:sec:multi_frame_infrared_small_target_detection}) and the medical image analysis application (Sec.~\ref{supp:sec:validation_on_medical_small_object_detection}).
Based on the proposed framework, we also plan to conduct more targeted analysis and exploration on more diverse visual perception tasks to further advance the overall community.

\parhead{Computational Efficiency.}
In our OPDC strategy described in Alg.~\ref{alg:matching}, we integrate the commonly-used assignment algorithm (\texttt{scipy.optimize.linear\_sum\_assignment}~\cite{linear_sum_assignment}) to find the minimum-cost matching between predicted targets and GT targets.
The algorithm ensures optimality but requires $O(n^3)$ time in worst-case scenarios.
While our experiments show real-time feasibility in different datasets, scaling to larger-scale targets may necessitate trade-offs between optimality and speed (\eg, via approximations) or hardware-specific optimizations.
Additionally, as a performance analysis framework, our method is primarily designed for offline evaluation of IRSTD applications, where runtime is not a critical concern.
However, to extend to application scenarios that may have real-time requirements, efficiency will also be one of the issues we will continue to focus on in the future.

\section{Societal Impacts}
\label{supp:societal_impacts}

This work rethinks the evaluation protocols in IRSTD, aiming to enhance research transparency and fairness through the hierarchical analysis framework and open-source benchmarking tool, ultimately supporting the development of more reliable systems for real-world deployment.
Furthermore, through systematic error analysis, this work identifies critical failure modes, which may help facilitate the exploration of necessary risk mitigation strategies in safety-sensitive applications.

\end{document}